%% file: active_learning_scm.tex
\documentclass{article}
\pdfoutput=1
 \usepackage[final]{nips_2017}

\usepackage[utf8]{inputenc} % allow utf-8 input
\usepackage[T1]{fontenc}    % use 8-bit T1 fonts
\usepackage{hyperref}       % hyperlinks
\usepackage{url}            % simple URL typesetting
\usepackage{booktabs}       % professional-quality tables
\usepackage{amsfonts}       % blackboard math symbols
\usepackage{nicefrac}       % compact symbols for 1/2, etc.
\usepackage{microtype}      % microtypography
\usepackage{xcolor}
\usepackage{enumitem}
\usepackage{wrapfig}

\usepackage{natbib}
\usepackage{multibib}
\usepackage{amsmath}
\usepackage{amsfonts}
\usepackage{amsthm}
\usepackage{amssymb}
\usepackage{mathtools}
\usepackage{algorithm,algorithmic}
\usepackage{subcaption}
\usepackage{tikz}
\usetikzlibrary{arrows}
\usetikzlibrary{positioning,automata,calc,patterns,decorations.pathmorphing,decorations.markings}

\DeclareMathOperator{\doop}{do}
\DeclareMathOperator{\pa}{\text{pa}}
\DeclareMathOperator{\var}{\text{var}}
\DeclareMathOperator{\val}{\text{val}}
\DeclareMathOperator{\nulli}{\varnothing}
\DeclareMathOperator*{\argmin}{arg\,min}

\newtheorem{example}{Example}
\newtheorem{definition}{Definition}
\newtheorem{lemma}{Lemma}

\usepackage{pgfplots}
\pgfplotsset{compat=newest}
\pgfplotsset{plot coordinates/math parser=false}
\pgfkeys{/pgfplots/mystyle/.style={
   semithick,
  xtick align = inside,
  ytick align = inside,
  ylabel shift = -2pt
  }}

\definecolor{color0}{rgb}{0.301960784313725,0.686274509803922,0.290196078431373}
\definecolor{color1}{rgb}{0.215686274509804,0.494117647058824,0.72156862745098}
\definecolor{color2}{rgb}{0.894117647058824,0.101960784313725,0.109803921568627}
\definecolor{color3}{RGB}{163, 141, 28}

\newlength{\figheight}
\newlength{\figwidth}

\tikzstyle{dashed} = [dash pattern=on 5pt off 1pt]

\title{Probabilistic Active Learning of Functions\\ in Structural Causal Models}

\author{
	Paul K.~Rubenstein$^{\ast}$, Ilya Tolstikhin, Philipp Hennig, Bernhard Sch\"olkopf\\
	Max Planck Institute for Intelligent Systems, T\"ubingen\\
	$^{\ast}$Machine Learning Group, University of Cambridge
}

\begin{document}

\maketitle

\begin{abstract}
  We consider the problem of learning the functions computing children from parents in a Structural Causal Model once the underlying causal graph has been identified. 
  This is in some sense the second step after causal discovery.
  Taking a probabilistic approach to estimating these functions, we derive a natural myopic active learning scheme that identifies the intervention which is optimally informative about all of the unknown functions jointly, given previously observed data. 
  We test the derived algorithms on simple examples, to demonstrate that they produce a structured exploration policy that significantly improves on unstructured base-lines.
%  Experiments on synthetic data confirm that actively choosing informative interventions speeds up learning compared to randomly intervening or learning only from observational data.
\end{abstract}
\vspace{-10pt}
\section{Introduction}
\label{sec:introduction}
\vspace{-5pt}
Large parts of the literature on causality are concerned with learning the causal graph of a system of random variables \citep{Spirtes2000,tong2001active, eberhardt2010causal, hyttinen2013experiment, mooij2014distinguishing}.
Also known as \emph{causal discovery} or \emph{causal inference}, this problem is motivated by realistic problems in science:
a biologist may wish to discover genes responsible for regulating other genes in a cell; 
a public health researcher may wish to know whether certain habits in a population (e.\,g.\ smoking) influence certain health outcomes (e.\,g.\ probability of developing lung cancer).

The starting point of this paper is to consider what should be done \emph{after} the causal graph of a system of variables has already been identified (be it by causal discovery methods, or from prior knowledge).
That is, it is known which variables are functions of which other variables, but the precise functional relationships are still unknown. 
Thus, although we understand the causal relationships in a coarse sense, we will not be able to accurately predict the result of an intervention to the system, possibly with implications for decision making.

For instance, suppose that in a cell, Gene A up-regulates Gene B and Gene B down-regulates Gene C. 
We know that reducing expression of Gene A will lead to a decrease in expression of Gene B which, in turn, will lead to an increase in expression of Gene C.
However, without more precise knowledge of the relationships between genes, we will be unable to quantitatively predict the effect of applying a drug that reduces expression of Gene A by 20\%.
Similarly, if our goal is to reduce overall levels of lung cancer in the population, then knowing only that smoking causes cancer is insufficient to know what the best public health policy should be: would it be better if we could persuade 50\% of smokers to stop smoking completely, or persuade every smoker to reduce their consumption by 50\%? 

For a \emph{passive} agent supplied with data generated by the system, learning the functional relationships between parents and children reduces simply to separate regression problems---one for each unknown function---once the causal graph is known.
Many knowledge acquisition problems, however, can be phrased as a sequential decision making process in which the data that is received at the next point in time is affected by a decision made based on the data that has already been observed.
A biologist does not blindly perform a series of costly and time-consuming experiments, only looking at the generated data once the last is over; the data from each experiment would be analysed before the next is performed, thus informing which experiment would be best to perform next. 

Below, we formalise this problem using the language of Structural Causal Models (SCMs), also known as Structural Equation Models \citep{Pearl2009,bollen2014structural}.
An SCM, in essence, consists of functions connecting child variables with their causal parents, and is equipped with a notion of \emph{intervention} in which a variable (or subset of variables) is externally forced to take a particular value (or values).
We use these interventions as an idealised mathematical representation of performing an experiment.
We take a Bayesian approach to estimating the functional relationships between parents and children, which naturally gives rise to an active learning algorithm to decide on the next `experiment' to perform.

\begin{figure}
	\begin{subfigure}{0.3\textwidth}
		\begin{tikzpicture}[->,>=stealth',auto,node distance=1.5cm,
		thick,main node/.style={circle,draw,font=\sffamily\bfseries}]
		\node[main node] (1) {$X_1$};
		\node[main node] (2) [right of=1] {$X_2$};
		\node[main node] (3) [right of=2] {$X_3$};
		
		\node (f2)  at ($(1)!0.5!(2)$) [yshift=0.5cm] {$f_2$};
		\node (f3)  at ($(2)!0.5!(3)$) [yshift=0.5cm] {$f_3$};
		
		\draw [->] (1) to (2);
		\draw [->] (2) to (3);
		\end{tikzpicture}
		\caption{\label{subfig:intro-graph-a}The observational setting}
	\end{subfigure}
	\hspace{0.03\textwidth}
	\begin{subfigure}{0.3\textwidth}
		\begin{tikzpicture}[->,>=stealth',auto,node distance=1.5cm,
		thick,main node/.style={circle,draw,font=\sffamily\bfseries}]
		\node[main node, style={fill=black,text=white}] (1) {$X_1$};
		\node[main node] (2) [right of=1] {$X_2$};
		\node[main node] (3) [right of=2] {$X_3$};
		
		\node (f2)  at ($(1)!0.5!(2)$) [yshift=0.5cm] {$f_2$};
		\node (f3)  at ($(2)!0.5!(3)$) [yshift=0.5cm] {$f_3$};
		
		\draw [->] (1) to (2);
		\draw [->] (2) to (3);
		\end{tikzpicture}
		\caption{\label{subfig:intro-graph-b}do($X_1=x_1$)}
	\end{subfigure}
	\hspace{0.03\textwidth}
	\begin{subfigure}{0.3\textwidth}
		\begin{tikzpicture}[->,>=stealth',auto,node distance=1.5cm,
		thick,main node/.style={circle,draw,font=\sffamily\bfseries}]
		\node[main node] (1) {$X_1$};
		\node[main node, style={fill=black,text=white}] (2) [right of=1] {$X_2$};
		\node[main node] (3) [right of=2] {$X_3$};
		
		\node (f3)  at ($(2)!0.5!(3)$) [yshift=0.5cm] {$f_3$};
		
		\draw [->] (2) to (3);
		\end{tikzpicture}
		\caption{\label{subfig:intro-graph-c}do($X_2=x_2$)}
	\end{subfigure}
	\setlength{\belowcaptionskip}{-10pt}
	\caption{\label{fig:intro-graph}Even in the simple setting of three variables whose graph is a chain, there is a non-trivial trade-off between the information one expects to gain by performing different interventions (see text).}
\end{figure}
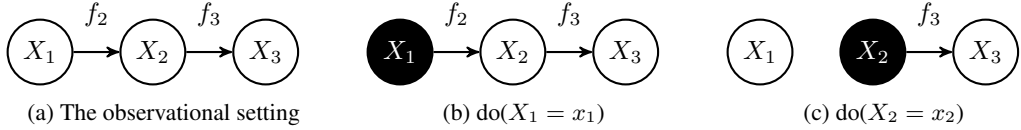

While our approach works with causal graphs that are arbitrary DAGs, the non-triviality of this problem is apparent even in the simple case of three variables whose causal graph is a chain (Figure~\ref{fig:intro-graph}).
Our goal in this situation is, in a sense to be made precise later, to learn the functions $f_2$ and $f_3$.
At each point in time, we must decide whether to perform one of the possible interventions or to passively observe the system, with each different action having some cost.
If we make a passive observation, we will learn \emph{something} about both $f_2$ and $f_3$, though only in areas where the distributions over $X_1$ and $X_2$ put probability mass. 
In the small sample setting, we are very unlikely to learn anything about the functions in areas of low probability of their inputs.
If we intervene on $X_1$ and choose what value to set it to, we can decide precisely \emph{where} we want to learn about $f_2$ and we will also learn about $f_3$ in some region, although we would be uncertain about where. 
If we intervene on $X_2$, we can learn a precise aspect of $f_3$, but will learn nothing about $f_2$.
How should we decide which action to pick, given what we already know about $f_2$ and $f_3$? 

The problem considered in this paper and our approach to solving it have a close connection to ideas in Bayesian optimization \citep{jones1998efficient, osborne2009gaussian, shahriari2016taking}. 
In Bayesian optimization, the goal is to find the extremum of a function that is expensive to evaluate, possibly exploiting known structure to speed up the search.
Due to this expense, the information obtained from each evaluation should be used efficiently.
In contrast, however, in our setting we are not simply interested in finding the extremum of an unknown function, but rather to learn the entire function (or set of functions) in some sense.
\citet{tong2000active} considered a similar setting, but only treated discrete variables.

Below, we begin with a formal definition of Structural Causal Models (Section~\ref{section:SCMs}). 
Section~\ref{section:problem-setup} states the precise problem we are tackling. Section~\ref{section:estimating-m-formalisation} provides a Bayesian formulation for inference in this setting, which is then complemented by an active learning scheme to guide exploration (Section~\ref{section:active-learning}). Section~\ref{section:experiments} provides empirical evaluations on synthetic toy examples.
\vspace{-10pt}
\section{Structural Causal Models}\label{section:SCMs}
\vspace{-10pt}
We now formally define Structural Causal Models (SCMs), the learning of which we will study in the remainder. For notational simplicity, our definition deviates from the wider literature by including the set of interventions modelled by the SCM.

\begin{definition}
Suppose that $X_1, \ldots, X_N, E_1,\ldots,E_N$ are variables with each $X_n$ and $E_n$ taking value in ${\mathcal{X}_n = \mathbb{R}}$ and $\mathcal{E}_n=\mathbb{R}$ respectively. 
We write ${X = (X_n)_{n=1,\ldots,N}}$ for the vector of variables, and ${\mathcal{X} = \prod_{n = 1,\ldots,N} \mathcal{X}_n}$ for their domain, similarly for $E$ and $\mathcal{E}$.
A Structural Causal Model (SCM) $\mathcal{M} = (\mathcal{S}, \mathcal{I}, \mathbb{P}_E)$ is a tuple consisting of the following three quantities:
\begin{itemize}[noitemsep, nolistsep]
	\item $\mathcal{S}$ is a set of \emph{structural equations} $X_n = f_n(X_{\pa(n)}, E_n)$, where $\pa(n)\subset\{1,\ldots,N\}$ and $X_{\pa(n)}$ is the vector of variables $(X_m)_{m\in\pa(n)}$. We call the $X_{\pa(n)}$ the \emph{causal parents} of $X_n$. 
	\item $\mathcal{I}$ is a set of \emph{interventions}. An intervention is a mathematical operation that replaces a subset of the equations in $\mathcal{S}$ with equations setting the variables to specific constants (e.\,g.\ $X_n=3$). 
	We write the intervention $i \in \mathcal{I}$ that intervenes on the subset of variables $X_{\var(i)}$, $\var(i) \subseteq \{1,\ldots,n\}$, setting them to values $x_{\val(i)}$, as $\doop(X_{\var(i)} = x_{\val(i)})$ (we will often abuse notation by writing $\doop(i)$ instead). We write $\mathcal{S}^{\doop(i)}$ for the resulting equations. 
	\item $\mathbb{P}_E$ is a distribution over the \emph{exogenous} (aka.~noise, unexplained) variables $E$ taking value in $\mathcal{E} = \mathbb{R}^N$. This distribution is fixed and does not change due to interventions.
\end{itemize}
\end{definition}
We will only consider acyclic\footnote{That is, the directed graph with nodes $\{1,\ldots,N\}$ with edges $n\rightarrow m$ if and only if $n\in \pa(m)$ is a DAG.} 
SCMs.
In this case, given any fixed value $e$ of the variables $E$, there is a unique value $x \in \mathcal{X}$ such that each structural equation in $\mathcal{S}$ is satisfied;\footnote{That is, for any $e\in\mathcal{E}$, there is a unique $x\in\mathcal{X}$ such that $x_n = f_n(x_{\pa(n)}, e_n)$ for each $n$. This can be seen to be true by explicitly writing each $x_n$ as a function of the $e_n$ by substituting the equations into one another.}
thus $\mathbb{P}_E$ induces a distribution over $\mathcal{X}$ via these unique solutions. 
We refer to this as the \emph{observational distribution} of the SCM, and write it as $\mathbb{P}_X^{\doop(\nulli)}$ (where $\nulli$ signifies the `empty' intervention). 
Under each intervention, the resulting intervened structural equations $\mathcal{S}^{\doop(i)}$ are also acyclic.
It follows by the same reasoning as above that the SCM implies a distribution $\mathbb{P}_X^{\doop(i)}$ over $\mathcal{X}$.
Therefore, once all of the parameters of the model $\mathcal{M}$ have been fixed, it implies a set of distributions indexed by intervention: $\{ \mathbb{P}_X^{\doop(i)} \: : \: i \in \mathcal{I} \}$. We assume that the empty intervention $\nulli$ is an element of $\mathcal{I}$.

An important note is that once the graphical structure has been fixed, the parameters of the model $\mathcal{M}$ are the functions $f_n$ and the distribution $\mathbb{P}_E$ over the variables $E$. 
We will make the non-trivial assumption that the structural equations are \emph{non-linear with additive Gaussian noise}, meaning that each equation is of the form $X_n = f_n(X_{\pa(n)}) + E_n$, where each $E_n$ is a zero-mean Gaussian random variable. This setting comes with results on the identifiability of the causal graph \citep{PetersMJS2011}.

\begin{example}
Consider the following SCM $\mathcal{M} = (\mathcal{S}, \mathcal{I}, \mathbb{P}_E)$ represented by Figure~\ref{fig:intro-graph}, where
\begin{minipage}{0.5\linewidth}
	\begin{align*}
	\mathcal{S} = \left\lbrace \begin{matrix*}[l]
	X_1 = 3 + E_1, \\ 
	X_2 = X_1^2 + E_2,\\ 
	X_3 = 2X_2 + \sin(X_2) + E_3
	\end{matrix*} \right\rbrace,
\end{align*}
\end{minipage}%
\begin{minipage}{0.5\linewidth}
	\begin{align*}
	&\mathcal{I} = \{ \nulli \} \cup \{\doop(X_i = x_i) \ | \  x_i \in \mathbb{R},\  i=1,2 \} ,\\
	&E_n \sim \mathcal{N}(0, \sigma_n^2), \quad \ n=1,2,3.
	\end{align*}
\end{minipage}
The observational distribution of $\mathcal{M}$ factorises as $\mathbb{P}_{X_1X_2X_3}^{\doop(\nulli)} = \mathbb{P}_{X_1}^{\doop(\nulli)}\mathbb{P}_{X_2|X_1}^{\doop(\nulli)}\mathbb{P}_{X_3|X_2}^{\doop(\nulli)}$. 
Under, for instance, the intervention $\doop(X_2=0)$, the equation $X_2=X_1^2 +E_2$ in $\mathcal{S}$ is replaced by $X_2=0$. Under the distribution $\mathbb{P}_{X_1X_2X_3}^{\doop(X_2=0)}$, $X_1$ and $X_3$ are independent and $X_2$ is degenerate.
\end{example}
\vspace{-10pt}
\section{Problem setup}
\label{section:problem-setup}
\vspace{-10pt}
Suppose  there exists an SCM ${\mathcal{M} = (\mathcal{S}, \mathcal{I}, \mathbb{P}_E)}$ with ${\mathcal{S} = \{X_n = f_n(X_{\pa(n)}) + E_n \: : \: n=1,\ldots,N\}}$ and $\mathbb{P}_E \sim \mathcal{N}(0, \Lambda)$ where $\Lambda$ is diagonal. 
We assume that the graphical structure is known, but the functions $f_n$ themselves are not.
For simplicity, we assume that $\Lambda$ is known.\footnote{This assumption could be relaxed to a more Bayesian approach involving a prior over the covariance matrix, which would reduce to running the algorithm derived in the following, but averaging its results over the posterior.}
We are given data $\mathcal{D}$ drawn from the observational and a variety of interventional distributions of $\mathcal{M}$; each element of $\mathcal{D}$ is a tuple $(i,x)$, where $x$ is an independent draw from $\mathbb{P}_X^{\doop(i)}$. We are interested in two separate tasks.

\textbf{Problem 1: Estimating $\mathcal{M}$.} Using $\mathcal{D}$, learn functions $\hat{f}_n$ such that the estimated model $\widehat{\mathcal{M}} = (\hat{\mathcal{S}}, \mathcal{I}, \mathbb{P}_E)$ with $\hat{\mathcal{S}} = \{X_n = \hat{f}_n(X_{\pa(n)}) + E_n \: : \: n=1,\ldots,N\}$ is `close' to the true model $\mathcal{M}$ in some sense. Part of this problem is to define a sensible notion of closeness between SCMs.

\textbf{Problem 2: Active learning.} We can select an intervention $i\in\mathcal{I}$ at some cost $c(i)$ and observe a single draw from $\mathbb{P}_X^{\doop(i)}$. Which $i$ should be selected to ensure the next estimation $\widehat{\mathcal{M}}$ after incorporating the new datum is as close to $\mathcal{M}$ as possible? 

The rest of this section is devoted to defining a risk functional to provide a notion of closeness between $\widehat{\mathcal{M}}$ and $\mathcal{M}$. There may be no single best way to define this notion of closeness, as desirable properties may be dependent on particular use case.\footnote{The same is true in the case of causal graph learning. See e.\,g. \cite{de2009comparison}.} 
Consider, for instance, the following scenarios in which the ultimate goal is to:
\begin{itemize}[noitemsep, nolistsep]
	\item Approximate each function $f_n$ so that a practitioner can visually interpret the relationship between parents and children (e.g., identifying genes as `excitatory' or `inhibitory', or identifying a threshold dosage at which a drug under medical trial is considered toxic.).
	\item Predict the result of an intervention that cannot be practically carried out, or is potentially dangerous to do so (e.g., raising interest rates, or giving a patient a drug).
	\item Better control a system in a variety of environments or conditions (e.g., learning the dynamics of a complex vehicle to be employed in variable conditions)
\end{itemize}
We will suppose that for each function $f_n$ with domain ${\mathcal{X}_{\pa(n)} = \prod_{m \in \pa(n)} \mathcal{X}_m}$ we are supplied with a probability measure $\Pi_n$ over $\mathcal{X}_{\pa(n)}$ specifying the importance of learning the pointwise value of $f_n$ at each input ${x_{\pa(n)}\in\mathcal{X}_{\pa(n)}}$. 
That is, $\Pi_n$ puts large amounts of mass in areas that we should learn $f_n$ precisely, small amounts of mass in areas that we should learn $f_n$ only approximately, and zero mass in areas for which we do not care about learning $f_n$ at all.
For the (estimated) function $\hat{f}_n$, we use the risk functional $L_n$ below. The weighted sum of these according to the importance of each function gives rise to the \emph{total risk} $L$, where $f=(f_n)_{n=1,\ldots,N}$ and $\hat{f}=(\hat{f}_n)_{n=1,\ldots,N}$ are vectors of the functions $f_n$ and $\hat{f}_n$, respectively.
\begin{align*}
L_n(\hat{f}_n || f_n) = \int_{\mathcal{X}_{\pa(n)}} \left( f_n(x) - \hat{f}_n(x) \right)^2 d\Pi_n(x), \qquad L(\hat{f} || f) = \sum_{n=1}^N \alpha_n L_n(\hat{f}_n || f_n), \quad \alpha_n \geq 0.
\end{align*}
We will assume for simplicity that $\alpha_n=1$ for each $n$.
$L_n$ is also known as the Mean Integrated Squared Error \citep{tsybakov2009introduction} and in the case that $\Pi_n = \mathbb{P}_{X_{\pa(n)}}^{\doop(\nulli)}$, this coincides with a typical objective that would be minimised in a classical non-parametric statistical learning setting \citep{gyorfi2006distribution}.
It is worth considering other possible risk functionals that could be used, since the presence of the measures $\Pi_n$ is arguably somewhat arbitrary in the $L$ that we consider.
When learning the parameters of a statistical model, a commonly used objective with many separate justifications is to minimise the KL divergence $\mathrm{KL}[\mathbb{P}_X || \hat{\mathbb{P}}_X ]$ between the true data distribution $\mathbb{P}_X$ and that implied by the learned model, $\hat{\mathbb{P}}_X$.
SCMs do not imply a single distribution over the variables X, but rather a family of distributions, one for each intervention: $\{ \mathbb{P}_X^{\doop(i)} \: : \: i \in \mathcal{I} \}$.
One may therefore wish to consider a separate loss for each interventional distribution and, for example, uniformly bound these losses over a subset of interventions $\mathcal{I}' \subseteq \mathcal{I}$ of interest.
\[ L_{\mathrm{KL}}^i (\hat{f}||f )  = \mathrm{KL}\left[\mathbb{P}^{\doop(i)}_X || \hat{\mathbb{P}}^{\doop(i)}_X \right] , \qquad\quad  L_{\mathrm{KL}}^{\mathcal{I}'} (\hat{f}||f )  = \sup_{i \in \mathcal{I}'} L_{\mathrm{KL}}^i (\hat{f}||f ).  \]
Alternatively, one could replace the KL divergence with a different divergence measure or metric on distributions. 
For instance, one could use the Maximum Mean Discrepancy (MMD) corresponding to a characteristic kernel $l$ \citep{5122}
\[ L_{\mathrm{MMD}_l}^i (\hat{f}||f )  = \mathrm{MMD}_l\left[\mathbb{P}^{\doop(i)}_X || \hat{\mathbb{P}}^{\doop(i)}_X \right] , \qquad\quad  L_{\mathrm{MMD}_l}^{\mathcal{I}'} (\hat{f}||f )  = \sup_{i \in \mathcal{I}'} L_{\mathrm{MMD}_l}^i (\hat{f}||f ).  \]
Though we do not analyse or derive active learning schemes for these risk functionals, they will be used in Section~\ref{section:experiments} to evaluate our algorithm. We leave their consideration for future work.
\vspace{-5pt}
\section{A probabilistic approach to learning $f$}\label{section:estimating-m-formalisation}
\vspace{-5pt}
By taking a Bayesian approach to learning the vector of unknown functions $f$, Problem 1 can be reduced to a series of independent regression problems between input and output domains $\mathcal{X}_{\pa(n)}$ and $\mathcal{X}_n$ for each $n$.
A common choice of prior when learning functions is a Gaussian Process (GP) \citep{Rasmussen:2005:GPM:1162254}.
For each function $f_n$, we will assume a zero mean GP prior with kernel $k_n$ over the domain $\mathcal{X}_{\pa(n)}$\footnote{No specific assumptions will be made on the choice of $k_n$, which can be freely chosen to incorporate prior knowledge about the functions (for instance, typical length scale of variation and magnitude). If $X_n$ is parentless, $f_n$ is an unknown \emph{constant} for which we assume a 1-dimensional Gaussian prior with zero mean and variance $k_n$.}
\[
f_n \sim \mathcal{GP}(0, k_n) .
\]
Recall that we are given a dataset $\mathcal{D}$ consisting of elements $(i,x)$ where $x\sim\mathbb{P}_X^{\doop(i)}$.
Let $\mathcal{D}_n$ be the collection of marginal observations of $(X_{\pa(n)},X_n)$ drawn from any distribution in which $X_n$ is not intervened upon.\footnote{That is, for any distribution $\mathbb{P}_X^{\doop(i)}$ for which $X_n \not \in \text{var}(i)$.}
Since by assumption $X_n \sim f_n(X_{\pa(n)}) + E_n$ where the distribution of $E_n \sim \mathcal{N}(0,\sigma_{n}^2)$ is known, each element $(x_{\pa(n)},x_n)$ of $\mathcal{D}_n$ represents an evaluation of $f_n$ at the input point $x_{\pa(n)}$ corrupted by Gaussian noise of known variance.
Performing GP regression using $\mathcal{D}_n$ as the data gives the posterior distribution over $f_n$.
By properties of Gaussians, this is also a GP with distribution
\[
f_n | \mathcal{D}_n \sim \mathcal{GP}(\mu_{f_n|\mathcal{D}_n} , k_{f_n|\mathcal{D}_n}),
\]
where $\mu_{f_n|\mathcal{D}_n}$ and $k_{f_n|\mathcal{D}_n}(x,y)$ can be explicitly written in terms of $k_n$ and the data $\mathcal{D}_n$ (see Appendix for details).
The above procedure can be applied for each $f_n$ independently, giving a posterior distribution over the vector of functions $f$.

\textbf{Which $\hat{f}$ should be chosen, given the posterior over $f$?} 
Problem 1 demands that a single choice  $\hat{f}$ be made when making the estimation $\widehat{\mathcal{M}}$ of $\mathcal{M}$.
For a fixed $\hat{f}$, the total risk $L(\hat{f}||f)$ is a random variable (the randomness coming from the uncertain belief over $f$).
The expectation of this random variable can be calculated and expressed in terms of the posterior covariance and mean functions of each $f_n$.
\begin{lemma}
	\begin{align*}
	\mathbb{E}_{f|\mathcal{D}} \left[ L(\hat{f} || f)\right] = \sum_{n=1}^{N} \alpha_n \int_{\mathcal{X}_{pa(n)}} \left(\hat{f}_n(x) - \mu_{f_n|\mathcal{D}_n}(x)\right)^2 +  k_{n|\mathcal{D}_n}(x,x) d\Pi_n(x)
	\end{align*}
\end{lemma}
See Appendix for proof. This immediately implies the following result.
\begin{lemma}
	Let $\mu_{f|\mathcal{D}}$ be the tuple of functions $(\mu_{f_n|\mathcal{D}_n})_{n=1,\ldots,N}$. Then
	\begin{align*}
	\mu_{f|\mathcal{D}} = \argmin_{\hat{f}} \mathbb{E}_{f | \mathcal{D}} \left[ L(\hat{f} || f)\right]
	\end{align*}
\end{lemma}
\vspace{-5pt}
	That is, choosing $\hat{f}_n$ to be the posterior mean of $f_n$ for each $n$ minimises the expected total risk.
	The uncertain distribution over $f$ directly yields an estimate of the total risk once the optimal $\hat{f} = \mu_{f|\mathcal{D}}$ is chosen which, once the prior has been fixed, is purely a function of the data $\mathcal{D}$.
Denote by ${\mathcal{R}(\mathcal{D}) = \mathbb{E}_{f | \mathcal{D}} \left[ L( \mu_{f|\mathcal{D}} || f)\right]}$ this \emph{expected total risk}.
Choosing the intervention $i$ for which $\mathcal{R}(\mathcal{D}\cup \{(i,x)\})$ is expected to be smallest after making the new observation $x$ from $\mathbb{P}_X^{\doop(i)}$ forms the basis of the proposed active learning algorithm.
\vspace{-5pt}
\section{Active learning}\label{section:active-learning}
\vspace{-5pt}
In this section a myopic active learning algorithm is derived based on the GP belief of the functions $f_n$ and the expected total risk $\mathcal{R}(\mathcal{D})$ described above.
At each step in time, we select an intervention $i$ at cost $c(i)$ and observe a single draw from the distribution $\mathbb{P}_X^{\doop(i)}$.
The goal is to select the intervention $i\in\mathcal{I}$ which will reduce the expected total risk as much as possible, taking into account the cost $c(i)$.
This problem is non-trivial for two main reasons.
\begin{enumerate}[noitemsep, nolistsep]
	\item The true distributions $\mathbb{P}_X^{\doop(i)}$ are unknown and therefore it is not possible to calculate the true expected reduction in expected total risk given a proposed intervention.
	\item There is a potentially large set of interventions that must be searched over.
\end{enumerate}
Consider the first issue above.
How will the expected total risk change if the intervention $i$ is chosen and a single new observation is drawn from $\mathbb{P}_X^{\doop(i)}$?
If the new observation is $x\in\mathcal{X}$, the new expected total risk will be $\mathcal{R}(\mathcal{D}\cup\{(i,x)\})$.
Define the \emph{value} of the intervention $i$ to be the expected reduction of $\mathcal{R}$ after performing the intervention $i$, divided by the cost of $i$:
\begin{equation}\label{eq:value}\tag{$\star$}
V(i | \mathcal{D}) =  \frac{\mathcal{R}(\mathcal{D} ) - \mathbb{E}_{x\sim \mathbb{P}^{\doop(i)}_X}\mathcal{R}\left(\mathcal{D}\cup \{(i,x)\} \right)}{c(i)}
\end{equation}
The goal is to find the intervention with the largest value, but since $\mathbb{P}_X^{\doop(i)}$ is unknown it is not possible to calculate the right-hand term in the numerator of (\ref{eq:value}).
It is possible, however, to \emph{estimate} this by replacing $\mathbb{P}_X^{\doop(i)}$ in the expectation with the \emph{belief} of the distribution based on the uncertain estimates of each $f_n$.
In this next two parts of this section, two different methods are proposed that estimate the expected total risk after performing each intervention. 
The first uses sampling, and requires a brute-force search over the set of interventions.
This may be appropriate when the set of possible interventions is small enough that this is feasible.
The second uses a form of dynamic programming, enabling a search over a larger set of interventions more efficiently.
The derived algorithm, however, makes specific assumptions on the graph of $\mathcal{M}$ and the set of interventions.
\begin{wrapfigure}[6]{R}{0.52\textwidth}
	\vspace{-50pt}
	\begin{minipage}{.52\textwidth}
		\begin{algorithm}[H]
			\algsetup{linenosize=\scriptsize}
			\caption{\small Sampling to estimate expected risk after intervention}
			\begin{algorithmic}[1]
				\scriptsize
				\STATE {\bf{Input:}} Previously observed data $\mathcal{D}$, GP kernels $k_n$ for prior on $f_n$, number of samples $T$, proposed intervention $i$.
				\FOR{$t=1,\ldots,T$}
				\STATE Draw $x^t$ from predictive distribution  $\widehat{\mathbb{P}}_X^{\doop(i)}$
				\STATE ${s_t = \mathcal{R}(\mathcal{D}\cup \{(i, x^t)\})}$: expected loss given new $x^t$
				\ENDFOR
				\RETURN $\frac{1}{T}\sum_{t=1}^T s_t$: estimated expected loss after intervention $i$.
			\end{algorithmic}
		\end{algorithm}
	\end{minipage}
\end{wrapfigure}

\subsection{Sampling}
Write $\widetilde{\mathbb{P}}_X^{\doop(i)}$ for the belief of $\mathbb{P}_X^{\doop(i)}$, taking into account the full uncertainty over $f$.\footnote{In contrast to $\widehat{\mathbb{P}}_X^{\doop(i)}$, which is the estimated distribution once a particular choice for $\hat{f}$ is made.}
Since the belief over each $f_n$ at each input $x_{\pa(n)}\in\mathcal{X}_{\pa(n)}$ is Gaussian and the noise variables are additive and Gaussian, it is possible to efficiently sample from $\widetilde{\mathbb{P}}_X^{\doop(i)}$.
It is illustrated here how to draw from the estimated observational distribution $\widetilde{\mathbb{P}}_X^{\doop(\nulli)}$ for notational convenience, but the procedure for any other $\widetilde{\mathbb{P}}_X^{\doop(i)}$ is essentially the same.

For any parentless variable, the structural equation is ${X_n = f_n + E_n}$ where ${f_n \sim \mathcal{N}(\mu_{n|\mathcal{D}_n}, k_{n|\mathcal{D}_n}) }$ and ${E_n \sim \mathcal{N}(0,\sigma_{n}^2)}$, and therefore ${X_n \sim \mathcal{N}(\mu_{n|\mathcal{D}_n}, k_{n|\mathcal{D}_n} + \sigma_{n}^2)}$.

For any variable with parents, the structural equation is ${X_n = f_n(X_{\pa(n)}) + E_n}$  where ${f_n \sim \mathcal{GP}(\mu_{n|\mathcal{D}_n}, k_{n|\mathcal{D}_n}) }$ and ${E_n \sim \mathcal{N}(0,\sigma_{n}^2)}$. Therefore the conditional distribution of a variable given its parents is ${X_n | X_{\pa(n)} \sim \mathcal{N}(\mu_{n|\mathcal{D}_n}(X_{\pa(n)}), k_{n|\mathcal{D}_n}(X_{\pa(n)},X_{\pa(n)}) + \sigma_{n}^2)}$.
Observe that the joint distribution $\widetilde{\mathbb{P}}_X^{\doop(\nulli)}$ factorises as
\begin{equation}\label{eq:decomposition}\tag{$\ast$}
\widetilde{\mathbb{P}}_X^{\doop(\nulli)} = \prod_{n : \pa(n) = \emptyset} \widetilde{\mathbb{P}}_{X_n}^{\doop(\nulli)}  \prod_{n:\pa(n)\not=\emptyset} \widetilde{\mathbb{P}}_{X_n|X_{\pa(n)}}^{\doop(\nulli)} .
\end{equation}
Since drawing from each of the above factors amounts to drawing from a Gaussian distribution, it is possible to efficiently sample from the entire joint distribution. By replacing ${\mathbb{P}}_X^{\doop(i)} $ with $\widetilde{\mathbb{P}}_X^{\doop(i)}$ in Equation~(\ref{eq:value}) above, we arrive at an estimate of the expected total risk which can be estimated using samples drawn from $\widetilde{\mathbb{P}}_X^{\doop(i)}$:
\begin{equation*}
\mathbb{E}_{x\sim \widetilde{\mathbb{P}}^{\doop(i)}_X}\mathcal{R}(\mathcal{D}\cup \{(i,x)\} )
\approx\frac{1}{T} \sum_{t=1}^T \mathcal{R}(\mathcal{D}\cup \{(i,x^t)\} )
\end{equation*}
where each $x^t\sim \widetilde{\mathbb{P}}^{\doop(i)}_X$ (see Algorithm 1). Finding the optimal intervention hence reduces to computing the above quantity for each $i\in\mathcal{I}$ from which it is possible to estimate each $V(i|\mathcal{D})$.

\begin{wrapfigure}[18]{r}{0.55\textwidth}
	\vspace{-48pt}
	\begin{minipage}{0.55\textwidth}
		\begin{algorithm}[H]
			\algsetup{linenosize=\scriptsize}
			\caption{\small Dynamic programming to estimate expected risk after interventions (chain, interventions on all $X_{n\leq m}$, some $m$ )}
			\begin{algorithmic}[1]
				\scriptsize
				\STATE {\bf{Input:}} Previously observed data $\mathcal{D}$, GP kernels $k_n$, discretisations $\widehat{\mathcal{X}}_n$ of each $\mathcal{X}_n$.
				\STATE	Pre-compute $U_n$ vectors and discrete approximations to conditional probability distributions:
				\FOR{$n=1,\ldots,N$}
				\STATE If $n=1$: $P^1_{x_{1}} \propto P(x_{1})$ for ${x_{1} \in \widehat{\mathcal{X}}_{1}}$, else:
				\STATE ${P^n_{x_{n-1},x_n} \propto P(x_{n}|x_{n-1})}$ for ${x_{n-1} \in \widehat{\mathcal{X}}_{n-1}}$, ${x_n \in \widehat{\mathcal{X}}_n}$
				\STATE  $U_n^{\text{curr}} = \mathcal{R}_n(\mathcal{D}_n)$
				\STATE  $U_n(x_{n-1}) $ for $x_{n-1}  \in \widehat{\mathcal{X}}_{n-1}$
				\ENDFOR
				\STATE Calculate new expected risk for all interventions:
				\FOR{$n=1,\ldots,N$}
				\STATE	$V = 0$
				\FOR{$m=N-1,\ldots,n+1$}
				\STATE$ V = P^m (V + U_{m+1})$
				\ENDFOR
				\STATE 	Expected risk after intervention on variables $X_m$, $m\leq n$:
				\STATE $ER_n =$ $V  + U_1^{\text{old}} +\ldots + U_n^{\text{old}}$
				\ENDFOR
				\RETURN Vectors $ER_n$  giving estimated expected risks after all interventions $\doop(X_n=x_n,X_{n-1}=\ldots)$
			\end{algorithmic}
		\end{algorithm}
	\end{minipage}
\end{wrapfigure}

\subsection{Dynamic programming}
If the set of interventions under consideration exhibits structure that coincides with that of the causal graph appropriately, it is possible to estimate the value of many interventions simultaneously.
A specific example is provided here of how this can be done in the case that the causal graph is a chain $X_1 \rightarrow \ldots \rightarrow X_N$, and any intervention intervenes on one variable and everything upstream of it.\footnote{That is, any intervention is of the form $\doop(X_n = x_n, \ n\leq m)$ for some $m$. This is equivalent to the case that all interventions act on a single variable, but only variables downstream of this are observable.}
A similar example for chains in which all interventions intervene on exactly one variable is provided in the Appendix.

The crux of this approach is the fact that the posterior covariance function of a Gaussian Process is only a function of the inputs of the conditioning data, not of the outputs.
That is, writing ${\mathcal{R}_n(\mathcal{D}_n) = \mathbb{E}_{f_n | \mathcal{D}_n} [L_n(\mu_{f_n|\mathcal{D}_n}, f_n)] = \int_{\mathcal{X}_{\pa(n)}} k_{n|\mathcal{D}_n} (x,x) d\Pi_n(x)}$ for the contribution to the expected total risk due to estimating function $f_n$, and writing ${\mathcal{D}_n = \{(x_{\pa(n)}^s, x_n^s) \, : \, s=1\,\ldots,|\mathcal{D}_n|\}}$, it follows that $\mathcal{R}_n$ is only actually a function of the $x_{pa(n)}^s$ (or, for parentless variables, just the size of the dataset $|\mathcal{D}_n|$).
Consider the intervention ${i = \doop({X_m=x_m},{X_{m-1}=\ldots})}$ that sets $X_m=x_m$ and all variables upstream of $X_m$ to arbitrary values. 
When a new observation  $x \sim \mathbb{P}_X^{\doop(i)}$ is made, this only provides new information about the functions $f_n$ for $n>m$, since $i$ intervenes on $X_n$ for $n\leq m$.
Let $U^{\text{curr}}_n = \mathcal{R}_n(\mathcal{D}_n)$ for $n \leq m$ be the current contributions to the expected total risk.
Define, for $n>m$, the following shorthand for the contribution to the new expected total risk function made by $f_n$ if a new observation of $f_n$ at the input point $x_{n-1}$ is made:
\begin{align*}
U_n(x_{n-1}) = \mathcal{R}_n(\mathcal{D}_n\cup \{(x_{n-1}, x_n )\}) 	\quad \quad \text{for any value } x_n, \  n > m .
\end{align*}
It follows that the estimated expected total risk after performing the intervention $i$ decomposes thus:
\begin{equation}\label{eq:dynamic}\tag{$\dagger$}
\mathbb{E}_{x\sim \widetilde{\mathbb{P}}^{\doop(i)}_X}\left[\mathcal{R}(\mathcal{D}\cup \{(i,x)\} ) \right]  =
\sum_{n=1}^m U^{\text{curr}}_n + 
\mathbb{E}_{x\sim \widetilde{\mathbb{P}}^{\doop(i)}_X} \left[ \sum_{n=m+1}^{N} U_n(x_{n-1}) \right] .
\end{equation}
By exploiting a factorisation of $\widetilde{\mathbb{P}}^{\doop(i)}_X$ similar to (\ref{eq:decomposition})  and discretely approximating the continuous domains $\mathcal{X}_n$, it is possible to reduce evaluating the right hand side of (\ref{eq:dynamic}) to a series of matrix multiplications and additions (see Appendix for details). 
This series of operations can be vectorised to allow calculation of (\ref{eq:dynamic}) for many $x_m$ simultaneously. 
Moreover, many of the initial calculations can be cached and used to speed up calculation over different values of $m$.
See Algorithm~2.
\section{Experiments}\label{section:experiments}

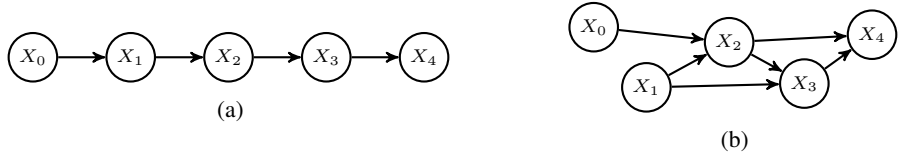
\begin{figure}
	\centering\scriptsize%
	\begin{subfigure}{0.48\textwidth}
		\centering
		\input{plots/ex1-graph-horizontal}
		\caption{\label{fig:graph-exp1}}
	\end{subfigure}%
	\begin{subfigure}{0.48\textwidth}
		\centering
		\input{plots/ex4-graph-horizontal}
		\caption{\label{fig:graph-exp4}}
	\end{subfigure}
	\caption{\label{fig:exp-graphs}The causal graphs of the SCMs (a) $\mathcal{M}_1$ and (b) $\mathcal{M}_2$ used in the experiments (Section~\ref{section:experiments}). }
\end{figure}

\begin{figure}
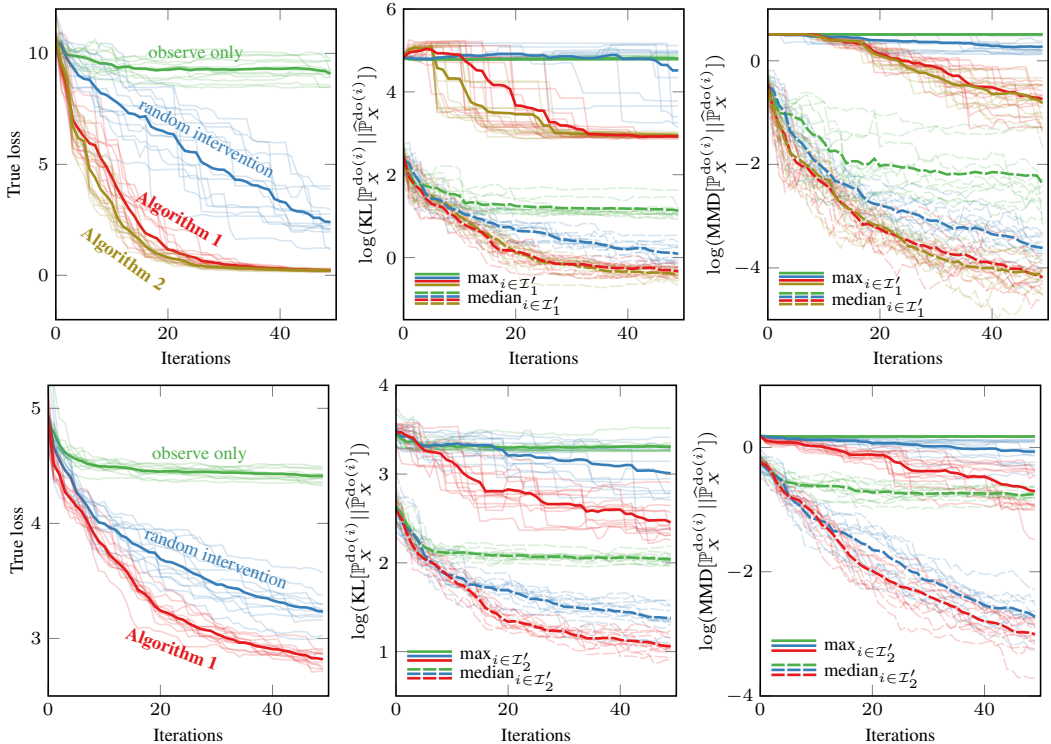

	\setlength{\figwidth}{.38\textwidth}
	\setlength{\figheight}{0.25\textheight}
	\centering\scriptsize%
		\input{plots/exp1-true-losses}
		\input{plots/exp1-kl-divergences}
		\input{plots/exp1-mmds}%
		\input{plots/exp4-true-losses}
		\input{plots/exp4-kl-divergences}
		\input{plots/exp4-mmds}
	\caption{\label{fig:results1}\textbf{Actively choosing informative interventions speeds up learning}. The proposed methods ({\color{color2}\textbf{Algorithm 1}}, {\color{color3}\textbf{Algorithm 2}}) outperform {\color{color0}\textbf{only observing}} and {\color{color1}\textbf{randomly intervening}} by each metric of evaluation. Top and bottom row show results of experiments on learning $\mathcal{M}_1$ and $\mathcal{M}_2$ respectively. Experimental details described in Section~\ref{section:experiments}. Each experiment was performed many times; faded lines represent results from single trials, bold lines represent averages of these single trials.}
\end{figure}

Figure~\ref{fig:results1} shows the results of running the proposed methods on two synthetic example SCMs $\mathcal{M}_1$ and $\mathcal{M}_2$, for which the graphs are given by Figures~\ref{fig:graph-exp1} and \ref{fig:graph-exp4} respectively.
In each case, structural equations consisting of sines and cosines of the parent variables were fixed, giving two sets of structural equations $\mathcal{S}_1$ and $\mathcal{S}_2$.
All noise variables in both SCMs were fixed to have variance $0.1$.
The interventions $\mathcal{I}_1$ for the chain SCM $\mathcal{M}_1$ were chosen to be all interventions of the form ${\doop(X_m=x_m,X_{m-1}=\ldots)}$ with ${x_m\in[-6,6]}$, such that $\mathcal{M}_1$ satisfies the conditions set out in the description of the dynamic programming algorithm.
The interventions $\mathcal{I}_2$ for the non-chain SCM $\mathcal{M}_2$ were defined to be all interventions on single variables $\doop(X_n=x_n)$ with $x_n \in [-6,6]$.
The total risk function for each experiment was chosen by setting $\Pi_n$ to be the uniform distribution on the domain $[-6,6]^{|\pa(n)|}$ and $\alpha_n=1$ for each $n$.
All GP kernels used were Radial Basis Function kernels with bandwidth parameter $1$. All costs $c(i)$ were assumed to be equal.

In each experiment, learning was initiated with no data.
For learning $\mathcal{M}_1$, the proposed algorithms were compared against the strategies of only drawing from the observational distribution of $\mathcal{M}_1$ and of selecting an intervention uniformly at random.
Since the dynamic programming algorithm could not be used for learning $\mathcal{M}_2$, we could only test the sampling algorithm.
Uniform discretisations $\mathcal{I}'_{1}$ and $\mathcal{I}'_{2}$ of the intervention sets were made of total size $250$ each.
These were used as the sets of interventions to search over using the sampling algorithm. 

The following quantities were used to evaluate the performance of the algorithms at each point in time:
the true total risk incurred by choosing $\hat{f}$ to be the vector of GP posterior means;
the maximum and median of the set of KL divergences $\mathrm{KL}[\mathbb{P}_X^{\doop(i)}|| \widehat{\mathbb{P}}_X^{\doop(i)}]$ calculated for each intervention $i\in \mathcal{I}'_{\ast}$ in the discretised intervention sets;
the maximum and median of the set of MMDs (corresponding to a Radial Basis Function kernel $l$ with bandwidth $1$) $\mathrm{MMD}_l[\mathbb{P}_X^{\doop(i)}|| \widehat{\mathbb{P}}_X^{\doop(i)}]$ calculated for each intervention $i\in \mathcal{I}'_{\ast}$ in the discretised intervention sets.
See Figure~\ref{fig:results1} for results.

\section{Discussion and future directions}\label{section:discussion}

There are many ways in which the work presented here could be incrementally furthered:
it may be possible to find efficient ways to search over the set of interventions subject to less restrictive assumptions than those made for Algorithm~2;
different distributions over the noise variables could be considered, in which case an approximation may need to be made  when regressing to find the posterior distribution over each $f_n$;
one could try to relax the additive noise assumption altogether.
Other natural extensions include estimating the value of an intervention based on reasoning multiple steps into the future, or considering the implications of a constrained budget.

Although the proposed algorithms seem to perform well on the synthetic toy examples considered, it remains to be seen whether this method, suitably extended, would similarly perform well on a convincing real-world problem. It is hard to find suitable real-world problems where convincing ground truth (for the functional relationships) exists, which is why we believe that it is sensible to assay such methods on synthetic data where performance can be accurately measured.

A perhaps more fundamental issue that was raised and \emph{not} tackled is the fact that it is not clear how best one should even define what it means to learn an SCM, or the parameters thereof.
We proposed $\sup_{i \in \mathcal{I}'} L_{\mathrm{KL}}^i (\hat{f}||f )$ and $\sup_{i \in \mathcal{I}'} L_{\mathrm{MMD}_l}^i (\hat{f}||f )$ for a suitable set of interventions $\mathcal{I}'$ as potentially more principled objectives to minimise than the total risk functional considered here.
Interestingly, the derived algorithms do reduce these quantities in the experiments considered, though this was in no way an explicit objective.
Future directions of research include trying to understand whether these, or other objectives, give rise to tractable methods for parameter estimation in causal models and for selecting interventions in active settings, and under what assumptions mathematical guarantees can be made.

\medskip

\small

\bibliography{active_learning_sem}

\newpage
\appendix
\section{Appendix}

\subsection{Posterior distribution of $f_n$ given $\mathcal{D}_n$}

The prior over $f_n$ is a Gaussian Process with zero mean and covariance function $k_n$
\[
f_n \sim \mathcal{GP}( 0 , k_n).
\]
Let $\mathcal{D}_n = \{(x_{\pa(n)}^s, x_n^s) \, : \, s=1\,\ldots,|\mathcal{D}_n|\}$ and let $K$ be the matrix with entries
\[
{K_{st} = k_i(x_{\pa(n)}^s,x_{\pa(n)}^t)}.
\]
Suppose that the Gaussian noise variable $E_n$ has variance $\sigma_n^2$. 
Then the posterior mean function $\mu_{f_n|\mathcal{D}_n}$ and posterior covariance function $k_{f_n|\mathcal{D}_n}$ can be written in closed form as
\begin{align*}
\mu_{f_n|\mathcal{D}_n} (x) &= \sum_{s,t=1}^{|\mathcal{D}_n|} k_n(x,x^s_{\pa(n)}) (K + \sigma_{n}^2)^{-1}_{st} x_n^t , \\
\\
k_{f_n|\mathcal{D}_n}(x,y) 
&=
k_n(x,y) - 
\sum_{s,t=1}^{|\mathcal{D}_n|} k_n(x,x_{\pa(n)}^s) (K+\sigma_{n}^2I)^{-1}_{st} k_n(y,x_{\pa(n)}^t) . \label{eqn:posterior-variance}
\end{align*}

\subsection{Proof of Lemma 1}
\begin{proof}
The expected total risk can be written in full form as
\begin{align*}
\mathbb{E}_{f|\mathcal{D}} \left[	L(\hat{f} || f)\right] &= \mathbb{E}_{f|\mathcal{D}} \left[	 \sum_{n=1}^{N} \alpha_n \int_{\mathcal{X}_{pa(n)}} \left(\hat{f}_n(x) - f_n(x)\right)^2 d\Pi_n(x) \right] \\
&= \sum_{n=1}^{N} \alpha_n \int_{\mathcal{X}_{pa(n)}}  \mathbb{E}_{f|\mathcal{D}} \left[	 \left(\hat{f}_n(x) - f_n(x)\right)^2 \right]  d\Pi_n(x) \\
&= \sum_{n=1}^{N} \alpha_n \int_{\mathcal{X}_{pa(n)}}  \mathbb{E}_{f_n|\mathcal{D}_n} \left[	 \left(\hat{f}_n(x) - f_n(x)\right)^2 \right]  d\Pi_n(x) . \\
\end{align*}
Under the posterior distribution given $\mathcal{D}_n$, we can decompose $f_n$ as the sum of its posterior mean and a zero mean Gaussian Process:
\begin{align*}
f_n | \mathcal{D}_n = \mu_{f_n|\mathcal{D}_n} + g_n, \qquad g_n \sim \mathcal{GP}(0, k_{f_n|\mathcal{D}_n}) .
\end{align*}
We can therefore rewrite the expected total risk as
\begin{align*}
&\mathbb{E}_{f|\mathcal{D}} \left[	L(\hat{f} || f)\right]
= \sum_{n=1}^{N} \alpha_n \int_{\mathcal{X}_{pa(n)}}  \mathbb{E}_{g_n|\mathcal{D}_n} \left[	 \left(\hat{f}_n(x) - \mu_{f_n|\mathcal{D}_n}(x) - g_n(x)\right)^2 \right]  d\Pi_n(x) \\
&= \sum_{n=1}^{N} \alpha_n  \int_{\mathcal{X}_{pa(n)}}  \mathbb{E}_{g_n|\mathcal{D}_n} \left[	 \left(\hat{f}_n(x) - \mu_{f_n|\mathcal{D}_n}(x)\right)^2  - \left(\hat{f}_n(x) - \mu_{f_n|\mathcal{D}_n}(x)\right) g_n(x) + g_n(x)^2 \right]  d\Pi_n(x) \\
&= \sum_{n=1}^{N} \alpha_n  \int_{\mathcal{X}_{pa(n)}}   \left(\hat{f}_n(x) - \mu_{f_n|\mathcal{D}_n}(x)\right)^2  - \left(\hat{f}_n(x) - \mu_{f_n|\mathcal{D}_n}(x)\right)\mathbb{E}_{g_n|\mathcal{D}_n} \left[  g_n(x) \right] + \mathbb{E}_{g_n|\mathcal{D}_n} \left[ g_n(x)^2 \right]  d\Pi_n(x) \\
&= \sum_{n=1}^{N} \alpha_n  \int_{\mathcal{X}_{pa(n)}}   \left(\hat{f}_n(x) - \mu_{f_n|\mathcal{D}_n}(x)\right)^2 + k_{f_n|\mathcal{D}_n}(x,x) d\Pi_n(x) . \\
\end{align*}
\end{proof}

\subsection{Derivation of Algorithm 2}
Under the intervention ${i = \doop({X_m=x^*_m},{X_{m-1}=\ldots})}$, the distribution $\widetilde{\mathbb{P}}_X^{\doop(i)}$ factorises as:
\begin{align*}
\widetilde{\mathbb{P}}_X^{\doop(i)} = \prod_{n \leq m } \delta_{X_n=x^*_n} \prod_{n\geq m+1} \widetilde{\mathbb{P}}_{X_n|X_{n-1}} ,
\end{align*}
where 
\begin{align*}
X_n|X_{n-1} &\sim \mathcal{N}\left(f_n(X_{n-1}), \sigma_n^2\right) \\
					&\sim \mathcal{N}\left(\mu_{f_n|\mathcal{D}_n}(X_{n-1}) , k_{f_n|\mathcal{D}_n}(X_{n-1},X_{n-1}) + \sigma_{n}^2 \right) .
\end{align*}
The quantity we are trying to calculate can hence be written
\begin{align*}
\mathbb{E}_{x\sim \widetilde{\mathbb{P}}^{\doop(i)}_X}\left[\mathcal{R}(\mathcal{D}\cup \{(i,x)\} ) \right] & =
\sum_{n=1}^m U^{\text{curr}}_n + 
\mathbb{E}_{x\sim \widetilde{\mathbb{P}}^{\doop(i)}_X} \left[ \sum_{n=m+1}^{N} U_n(x_{n-1}) \right] \\
& = \sum_{n=1}^m U^{\text{curr}}_n +  \int_{\mathcal{X}} \sum_{n=m+1}^{N} U_n(x_{n-1})  \ d\widetilde{\mathbb{P}}^{\doop(i)}_X(x) \\
& = \sum_{n=1}^m U^{\text{curr}}_n +   \sum_{n=m}^{N-1} \int_{\mathcal{X}_n} U_{n+1}(x_{n})  \ d\widetilde{\mathbb{P}}_{X_n}^{\doop(i)}(x_n) . \\
\end{align*}
Now, since $\widetilde{\mathbb{P}}_{X_{m}}^{\doop(i)} = \delta_{X_m=x^*_m}$, observe that we can write
\begin{align*}
	 \int_{\mathcal{X}_m} U_{m+1}(x_{m})  \ d\widetilde{\mathbb{P}}_{X_m}^{\doop(i)}(x_m) &= U_{m+1}(x^*_{m}) .\\
\end{align*}
For any $n>m$, we have
\begin{align*}
\int_{\mathcal{X}_{n}} U_{n+1}(x_{n}) &  \ d\widetilde{\mathbb{P}}_{X_{n}}^{\doop(i)}(x_{n}) \\
&= \int_{\mathcal{X}_{m}} \ldots \int_{\mathcal{X}_{n}} U_{n+1}(x_{n})  \ d\widetilde{\mathbb{P}}_{X_{n} | X_{n-1}}^{\doop(i)}(x_{n}|x_{n-1}) \ldots d\widetilde{\mathbb{P}}_{X_{m+1}|X_m}^{\doop(i)}(x_{m+1}|x_{m}) d\widetilde{\mathbb{P}}_{X_{m}}^{\doop(i)}(x_{m}) . 
\end{align*}
Hence if we define the following quantities recursively
\begin{align*}
V_{N-1}(x_{N-2}) &= \int_{\mathcal{X}_{N-1}} U_{N}(x_{N-1}) \ 	 d\widetilde{\mathbb{P}}_{X_{N-1}|X_{N-2}}^{\doop(i)}(x_{N-1}|x_{N-2}), \\
V_{n}(x_{n-1}) &= \int_{\mathcal{X}_{n}} V_{n+1}(x_{n}) + U_{n+1}(x_{n}) \ 	 d\widetilde{\mathbb{P}}_{X_{n}|X_{n-1}}^{\doop(i)}(x_{n}|x_{n-1}) \qquad n=N-2,\ldots m+1 ,
\end{align*}
it follows that 
\[
V_{m+1}(x^*_m) = \sum_{n=m}^{N-1} \int_{\mathcal{X}_n} U_{n+1}(x_{n})  \ d\widetilde{\mathbb{P}}_{X_n}^{\doop(i)}(x_n) .
\]
We can approximate calculation of $V_{m+1}(x^*_m)$ for many $x^*_m$ simultaneously by discretising each $\mathcal{X}_n$ into a set of points $x_n^1, x_n^2, \ldots, x_n^{D_n}$.
Define $P^n$ to be the matrix with normalised rows such that ${P^n_{ij} \propto \ \widetilde{p}_{X_{n}|X_{n-1}}^{\doop(i)}(x^j_{n}|x^i_{n-1})}$, where $\widetilde{p}^{\doop(i)}$ is the density of $\widetilde{\mathbb{P}}^{\doop(i)}$ with respect to the Lebesgue measure on $\mathcal{X}$.
Define $\mathbf{u}^n$ to be the vector with entries $\mathbf{u}^n_i = U_n(x_{n-1}^i)$.
Then, if we recursively define
\begin{align*}
\mathbf{v}^{N-1}  &= P^{N-1} \mathbf{u}_{N},  \\
\mathbf{v}^{n} &= P^n (\mathbf{v}^{n+1} + \mathbf{u}_{n+1} )\qquad n=N-2,\ldots m+1 ,
\end{align*}

it follows that $\mathbf{v}^{m+1}_i \approx V_{m+1}(x_m^i)$.

\subsection{Another Dynamic Programming Algorithm: chain, single variable interventions.}

Similar reasoning to the above can be used to derive a dynamic programming scheme to calculate the estimated total risk after a proposed intervention $i = \doop(X_m = x^*_m)$ (i.\,e.\ intervening on a \emph{single} variable) for many $x_m^*$ simultaneously.
This is summarised by Algorithm~3.

Under this intervention, the joint distribution factorises as
\begin{align*}
\widetilde{\mathbb{P}}_X^{\doop(i)} =  \widetilde{\mathbb{P}}_{X_1}\prod_{n = 2}^{m-1} \widetilde{\mathbb{P}}_{X_n|X_{n-1}} \delta_{X_m=x^*_m} \prod_{n = m+1}^{N} \widetilde{\mathbb{P}}_{X_n|X_{n-1}} .
\end{align*}
When we intervene on a single variable $X_m$, we learn something new about all functions except $f_m$. 
We can write the expected new total risk, the quantity we want to evaluate, as
\begin{align*}
\mathbb{E}_{x\sim \widetilde{\mathbb{P}}^{\doop(i)}_X}\left[\mathcal{R}(\mathcal{D}\cup \{(i,x)\} ) \right] & =
U_1 + \mathbb{E}_{x\sim \widetilde{\mathbb{P}}^{\doop(i)}_X} \left[ \sum_{n=2}^{m-1} U_n(x_{n-1}) \right] +
U^{\text{curr}}_m + 
\mathbb{E}_{x\sim \widetilde{\mathbb{P}}^{\doop(i)}_X} \left[ \sum_{n=m+1}^{N} U_n(x_{n-1}) \right] .\\
\end{align*}
Each of the expectations above can be calculated recursively in a similar fashion to the strategy employed above. If we define
\begin{align*}
V_{N-1}(x_{N-2}) &= \int_{\mathcal{X}_{N-1}} U_{N}(x_{N-1}) \ 	 d\widetilde{\mathbb{P}}_{X_{N-1}|X_{N-2}}^{\doop(i)}(x_{N-1}|x_{N-2}) , \\
V_{n}(x_{n-1}) &= \int_{\mathcal{X}_{n}} V_{n+1}(x_{n}) + U_{n+1}(x_{n}) \ 	 d\widetilde{\mathbb{P}}_{X_{n}|X_{n-1}}^{\doop(i)}(x_{n}|x_{n-1}), \qquad n=N-2,\ldots m+1, 
\end{align*}
it follows that 
\[
V_{m+1}(x^*_m) = \mathbb{E}_{x\sim \widetilde{\mathbb{P}}^{\doop(i)}_X} \left[ \sum_{n=m+1}^{N} U_n(x_{n-1}) \right] .
\]
Similarly, defining
\begin{align*}
V_{m-2}(x_{m-3}) &= \int_{\mathcal{X}_{m-2}} U_{m-1}(x_{m-2}) \ 	 d\widetilde{\mathbb{P}}_{X_{m-2}|X_{m-3}}^{\doop(i)}(x_{m-2}|x_{m-3}), \\
V_{n}(x_{n-1}) &= \int_{\mathcal{X}_{n}} V_{n+1}(x_{n}) + U_{n+1}(x_{n}) \ 	 d\widetilde{\mathbb{P}}_{X_{n}|X_{n-1}}^{\doop(i)}(x_{n}|x_{n-1}), \qquad n=m-3,\ldots 2, \\
V_{1} &= \int_{\mathcal{X}_{1}} V_{2}(x_{1}) + U_{2}(x_{1}) \ 	 d\widetilde{\mathbb{P}}_{X_{1}}^{\doop(i)}(x_{1}), \\
\end{align*}
it follows that 
\[
V_{1} = \mathbb{E}_{x\sim \widetilde{\mathbb{P}}^{\doop(i)}_X} \left[ \sum_{n=2}^{m-1} U_n(x_{n-1}) \right] . 
\]
As before, we can approximate calculation of $V_{m+1}(x^*_m)$ for many $x^*_m$ simultaneously by discretising each $\mathcal{X}_n$ into a set of points $x_n^1, x_n^2, \ldots, x_n^{D_n}$.
Define $P^n$ to be the matrix with normalised rows such that ${P^n_{ij} \propto \ \widetilde{p}_{X_{n}|X_{n-1}}^{\doop(i)}(x^j_{n}|x^i_{n-1})}$  for $n>1$, where $\widetilde{p}^{\doop(i)}$ is the density of $\widetilde{\mathbb{P}}^{\doop(i)}$ with respect to the Lebesgue measure on $\mathcal{X}$. 
Define $P^1$ to be the normalised \emph{vector} with $P^1_i \propto \widetilde{p}_{X_1}(x^i_1)$.
Define $\mathbf{u}^n$ to be the vector with entries $\mathbf{u}^n_i = U_n(x_{n-1}^i)$.
Then, if we recursively define
\begin{align*}
\mathbf{v}^{N-1}  &= P^{N-1} \mathbf{u}_{N}, \\
\mathbf{v}^{n} &= P^n (\mathbf{v}^{n+1} + \mathbf{u}_{n+1} )\qquad n=N-2,\ldots m+1,
\end{align*}
it follows that $\mathbf{v}^{m+1}_i \approx V_{m+1}(x_m^i)$. 
We must also estimate $V_1$.
If we define
\begin{align*}
\mathbf{v}^{m-2}  &= P^{m-2} \mathbf{u}_{m-2}, \\
\mathbf{v}^{n} &= P^n (\mathbf{v}^{n+1} + \mathbf{u}_{n+1} )\qquad n=m-3,\ldots 2,\\
\mathbf{v}^{1} &= {P^1}^{\intercal} (\mathbf{v}^{2} + \mathbf{u}_{2} )
\end{align*}
then it follows that $\mathbf{v}^1 \approx V_1$.

\begin{figure}
		\begin{algorithm}[H]
	\algsetup{linenosize=\scriptsize}
			\caption{\small Dynamic programming to estimate expected risk after interventions (chain, interventions on single variable $X_{m}$)}
			\begin{algorithmic}[1]
				\scriptsize
				\STATE {\bf{Input:}} Previously observed data $\mathcal{D}$, GP kernels $k_n$, discretisations $\widehat{\mathcal{X}}_n$ of each $\mathcal{X}_n$.
				\STATE	Pre-compute $U_n$ vectors and discrete approximations to conditional probability distributions:
				\FOR{$n=1,\ldots,N$}
					\STATE ${P^n_{x_{n-1},x_n} \propto P(x_{n}|x_{n-1})}$ for ${x_{n-1} \in \widehat{\mathcal{X}}_{n-1}}$, ${x_n \in \widehat{\mathcal{X}}_n}$
					\STATE  $U_n^{\text{curr}} = \mathcal{R}_n(\mathcal{D}_n)$
					\STATE  $U_n(x_{n-1}) $ for $x_{n-1}  \in \widehat{\mathcal{X}}_{n-1}$
				\ENDFOR
				\STATE Calculate expected loss for all interventions on each variable in turn:
				\FOR{$m=1,\ldots,N$}
				\STATE	$V = 0$
				\FOR{$n=N-1,\ldots,m+1$}
				\STATE$ V = P^n (V + U_n)$ 
				\ENDFOR
				\STATE	$V' = 0$
				\FOR{$n=m-1,\ldots,2$}
				\STATE$ V' = P^n (V' + U_n)$ 
				\ENDFOR
				\STATE 	Expected risk after intervention on variable $m$: $ER_m =$ $V + V' + U_1 + U_m^{\text{curr}}$ 
				\ENDFOR
				\RETURN Vectors $ER_n$  giving estimated expected risks after interventions for all interventions on $X_n$.
			\end{algorithmic}
		\end{algorithm}
\end{figure}

\end{document}

%% file: plots/ex1-graph-horizontal.tex
\begin{tikzpicture}[->,>=stealth',auto,node distance=1.3cm,
		thick,main node/.style={circle,draw,font=\sffamily\bfseries}]
		\node[main node] (0) {$X_0$};
		\node[main node] (1) [right of=0]{$X_1$};
		\node[main node] (2) [right of=1] {$X_2$};
		\node[main node] (3) [right of=2] {$X_3$};
		\node[main node] (4) [right of=3] {$X_4$};

%		\node (f1)  at ($(0)!0.5!(1)$) [xshift=0.5cm] {$f_1$};
%		\node (f2)  at ($(1)!0.5!(2)$) [xshift=0.5cm] {$f_2$};
%		\node (f3)  at ($(2)!0.5!(3)$) [xshift=0.5cm] {$f_3$};
%		\node (f4)  at ($(3)!0.5!(4)$) [xshift=0.5cm] {$f_4$};

		\draw [->] (0) to (1);		
		\draw [->] (1) to (2);
		\draw [->] (2) to (3);
		\draw [->] (3) to (4);
\end{tikzpicture}

%% file: plots/ex4-graph-horizontal.tex
\begin{tikzpicture}[->,>=stealth',auto,node distance=1.3cm,
		thick,main node/.style={circle,draw,font=\sffamily\bfseries}]
		\node[main node] (0) {$X_0$};
		\node[main node] (1) [below of=0, yshift=0.50cm, xshift=0.7cm]{$X_1$};
		\node[main node] (2) [right of=0, yshift=-0.2cm, xshift=0.5cm] {$X_2$};
		\node[main node] (3) [right of=1, yshift=0.05cm, xshift=0.8cm] {$X_3$};
		\node[main node] (4) [right of=0, yshift=-0.1cm, xshift=2.4cm] {$X_4$};

%		\node (f1)  at ($(0)!0.5!(1)$) [xshift=0.5cm] {$f_1$};
%		\node (f2)  at ($(1)!0.5!(2)$) [xshift=0.5cm] {$f_2$};
%		\node (f3)  at ($(2)!0.5!(3)$) [xshift=0.5cm] {$f_3$};
%		\node (f4)  at ($(3)!0.5!(4)$) [xshift=0.5cm] {$f_4$};

		\draw [->] (0) to (2);		
		\draw [->] (1) to (2);
		\draw [->] (1) to (3);
		\draw [->] (2) to (3);
		\draw [->] (2) to (4);
		\draw [->] (3) to (4);
\end{tikzpicture}

%% file: plots/exp4-true-losses.tex
% This file was created by matplotlib2tikz v0.6.7.
\begin{tikzpicture}

\definecolor{color1}{rgb}{0.215686274509804,0.494117647058824,0.72156862745098}
\definecolor{color0}{rgb}{0.301960784313725,0.686274509803922,0.290196078431373}
\definecolor{color2}{rgb}{0.894117647058824,0.101960784313725,0.109803921568627}

\begin{axis}[
xmin=0, xmax=50,
ymin=2.5, ymax=5.2,
axis on top,
width=\figwidth,
height=\figheight,
tick pos=both,
xlabel={Iterations},
ylabel={True loss},
mystyle
]
\addplot [color0, opacity=0.2]
table {%
0 4.92424720981218
1 4.79493618250434
2 4.62761195393966
3 4.59995782125584
4 4.59101460482371
5 4.56907773281756
6 4.57289114022147
7 4.52619167804189
8 4.50824087725564
9 4.49547239950373
10 4.46126574558227
11 4.45481078251406
12 4.44586541612242
13 4.42843709235167
14 4.42476728217436
15 4.41802905852452
16 4.41382331569551
17 4.42359570143714
18 4.43248665383186
19 4.43249042302029
20 4.44448633823853
21 4.44339862525362
22 4.44666926236789
23 4.45981281895764
24 4.48110331248188
25 4.46974531536014
26 4.47430625242249
27 4.47102462459631
28 4.47539100295042
29 4.46063098903428
30 4.46190709708335
31 4.46126275515397
32 4.4580618784199
33 4.45597898403234
34 4.45357017899021
35 4.45676815592251
36 4.46391853426415
37 4.43491536340218
38 4.44153557151086
39 4.44185104633032
40 4.43547663682248
41 4.43162959224457
42 4.43133034955909
43 4.43317563363016
44 4.43680970383598
45 4.42697631298612
46 4.42034301481434
47 4.42254217650233
48 4.43092390401812
49 4.43045736440513
};
\addplot [color0, opacity=0.2]
table {%
0 4.9972696842653
1 4.80193527129404
2 4.66690315732485
3 4.65426059787377
4 4.67912687470898
5 4.63376747815765
6 4.56620417472564
7 4.55213610228326
8 4.54176728649333
9 4.54085509078304
10 4.55493951388508
11 4.54695366580299
12 4.54433356833042
13 4.51020721193176
14 4.52352636419445
15 4.53385006854691
16 4.52034576110385
17 4.5217533585262
18 4.51988490838304
19 4.51463680964177
20 4.50438721656547
21 4.4891633850425
22 4.49195720903053
23 4.49171833868842
24 4.48285359775302
25 4.47912102856208
26 4.50411088375495
27 4.51311026306755
28 4.5115922379157
29 4.49393929126752
30 4.4994339072347
31 4.46942820984268
32 4.46557987257647
33 4.46002497656827
34 4.46408531832
35 4.46603466190304
36 4.47835580697417
37 4.4764233989276
38 4.48223087483502
39 4.48437186623105
40 4.48013648135876
41 4.4820085733875
42 4.48786554354608
43 4.48743208824034
44 4.48621482559395
45 4.48296568766624
46 4.47686104774306
47 4.4748778381446
48 4.46832617643223
49 4.46941542169037
};
\addplot [color0, opacity=0.2]
table {%
0 4.692182367182
1 4.66043822927545
2 4.63761493326718
3 4.59341835689535
4 4.48969069189878
5 4.49674596086673
6 4.46413752849642
7 4.47479483115973
8 4.49225133437346
9 4.49241812613737
10 4.47105737958777
11 4.45780097829853
12 4.46446039396347
13 4.47873960113007
14 4.4779136058613
15 4.40551947888221
16 4.40108636129634
17 4.38472533158535
18 4.3958261189937
19 4.38175630335001
20 4.36440287819173
21 4.3731883502292
22 4.36101207739523
23 4.35871855886601
24 4.36519495001373
25 4.36601038308032
26 4.36671844639331
27 4.3670769004053
28 4.37401600845636
29 4.35840622547802
30 4.35518074873383
31 4.35529101162198
32 4.35982743865877
33 4.36627689970023
34 4.35718104779429
35 4.35944980537067
36 4.35791644818886
37 4.35755594755556
38 4.35284616823016
39 4.35191442717874
40 4.34915480687634
41 4.35160701038976
42 4.35049566414601
43 4.35026768261259
44 4.3568415659084
45 4.35870767085638
46 4.3623394454661
47 4.36128595668804
48 4.36170139884561
49 4.36268204326523
};
\addplot [color0, opacity=0.2]
table {%
0 4.92310312641728
1 4.87300646082976
2 4.67337255428837
3 4.60490216871819
4 4.60173787166773
5 4.58849035960864
6 4.55223052050431
7 4.5412207279743
8 4.52105660735098
9 4.51094489793968
10 4.52172012866689
11 4.52375383801398
12 4.54772649087337
13 4.56363578329613
14 4.55224355444884
15 4.54036386247855
16 4.51981770738922
17 4.50475345091816
18 4.50054060800607
19 4.49903228026608
20 4.50166069394431
21 4.51287764216573
22 4.51065918427884
23 4.51610713344723
24 4.5138911854902
25 4.52039929000652
26 4.52690888556117
27 4.52382268511513
28 4.50123822767825
29 4.5031795455391
30 4.4960464540738
31 4.49535669921664
32 4.4885606352808
33 4.486110400536
34 4.47897453224781
35 4.45033633831422
36 4.4411928789031
37 4.43704703434216
38 4.43539043310046
39 4.42465739075093
40 4.42709268054137
41 4.42917510827471
42 4.42878334766487
43 4.42606686967374
44 4.42526824430849
45 4.41471718597498
46 4.41365194581409
47 4.41439501892617
48 4.41689241057908
49 4.41966596131664
};
\addplot [color0, opacity=0.2]
table {%
0 4.84184101459916
1 4.86483648914625
2 4.69709595124943
3 4.55558755769914
4 4.59824922167481
5 4.59596474955333
6 4.63534442669933
7 4.61801822724777
8 4.57509600965508
9 4.57435654769281
10 4.55838226243438
11 4.55522558604892
12 4.56154284531507
13 4.51648055520428
14 4.5037770742906
15 4.50121110075375
16 4.51430852047574
17 4.52124504675594
18 4.52680241647274
19 4.52144889535281
20 4.52091759199705
21 4.52091183587619
22 4.52003461718591
23 4.49916261778051
24 4.4790177516649
25 4.47196978138128
26 4.4675574381187
27 4.46463250674121
28 4.46201423845535
29 4.46284908797892
30 4.46397564956039
31 4.46215929169554
32 4.46098125014995
33 4.46161350535371
34 4.45718432582549
35 4.4560494070381
36 4.4590766616935
37 4.44957411089638
38 4.44614308634066
39 4.44296696472238
40 4.43759883064289
41 4.43853123206586
42 4.43865105416023
43 4.44857258752455
44 4.44972117734549
45 4.44880796557895
46 4.42893219592201
47 4.4326421423309
48 4.43046480389124
49 4.41656262036185
};
\addplot [color0, opacity=0.2]
table {%
0 4.96270562895296
1 4.69831291818387
2 4.64123980174533
3 4.58827057458055
4 4.54715838540916
5 4.5221687772482
6 4.55474842754323
7 4.5524630761611
8 4.55396494946092
9 4.55430484905854
10 4.569291185824
11 4.51917094551464
12 4.51666000125122
13 4.52164850744146
14 4.51299245882168
15 4.51300841258475
16 4.50770390461913
17 4.46244210636873
18 4.45988842893622
19 4.4473570742062
20 4.44648195086344
21 4.4403592754867
22 4.44124007175593
23 4.44097028678286
24 4.44343634022453
25 4.44413818415764
26 4.44885047763062
27 4.44703196203222
28 4.45348999150307
29 4.46380824665066
30 4.49067664354348
31 4.48974051720866
32 4.49407128130545
33 4.48517410821789
34 4.50167413746219
35 4.50279791345242
36 4.500109490696
37 4.51614730112263
38 4.51685072206351
39 4.51530001379948
40 4.50906768490014
41 4.51107423836239
42 4.51304499888242
43 4.51366405694614
44 4.51292455595606
45 4.5110327653004
46 4.510129344231
47 4.49819871762652
48 4.49655884307821
49 4.4990025749024
};
\addplot [color0, opacity=0.2]
table {%
0 5.45016237482302
1 5.20322035481412
2 4.916163332068
3 4.87102818531169
4 4.66052653596533
5 4.64863174972276
6 4.65957277189121
7 4.52441248138827
8 4.53283141328719
9 4.49161741343476
10 4.50842324909561
11 4.52770188653353
12 4.52844067947016
13 4.50335875487577
14 4.49399613828901
15 4.46863333882298
16 4.44370914715653
17 4.43552945939329
18 4.4317314248446
19 4.43594981902184
20 4.42430172856445
21 4.42294238283318
22 4.4328158351988
23 4.43309922163244
24 4.43063391906551
25 4.41745068488641
26 4.41429817969929
27 4.41282387854771
28 4.41163157622917
29 4.40573546273247
30 4.39234189491236
31 4.39561735690835
32 4.39176832072823
33 4.38383288067258
34 4.38350418599012
35 4.38628637208571
36 4.38625121144107
37 4.38795969863255
38 4.38831926047048
39 4.38975809749089
40 4.39028770896562
41 4.39814695659763
42 4.39911831481742
43 4.39969243162397
44 4.39976946682799
45 4.39788262580838
46 4.38903907003085
47 4.39159367927261
48 4.39694915669201
49 4.39817682406893
};
\addplot [color0, opacity=0.2]
table {%
0 4.91530506901616
1 4.84216154063171
2 4.72271067146403
3 4.72012162965146
4 4.59306210456346
5 4.56199410908119
6 4.49553134345087
7 4.49412331643877
8 4.49076458666182
9 4.47741710554672
10 4.47001819514946
11 4.47616480933408
12 4.4750621314615
13 4.48526727683119
14 4.48669945332183
15 4.48458046283407
16 4.44044226964915
17 4.43522017265723
18 4.44243028204757
19 4.44478816535818
20 4.44931989840775
21 4.48368477315025
22 4.4701465018231
23 4.46857199101362
24 4.46648144433624
25 4.47880551565134
26 4.47679960966639
27 4.47162727363595
28 4.46782694019304
29 4.4875258227368
30 4.47712328166225
31 4.4796365817543
32 4.49186892310847
33 4.49598772436134
34 4.49009539319697
35 4.49487446784242
36 4.49373236104854
37 4.49311433212585
38 4.49230250031422
39 4.50380276009147
40 4.50651889672729
41 4.50873871532614
42 4.50458271956239
43 4.48372330875708
44 4.4845847595234
45 4.48493424565745
46 4.48513798922656
47 4.48392007188898
48 4.48490384269641
49 4.4805570039903
};
\addplot [color0, opacity=0.2]
table {%
0 4.68154143385065
1 4.71725675201939
2 4.72960294317321
3 4.56585315188566
4 4.55321497998181
5 4.49655383534689
6 4.50228522609197
7 4.48464481094566
8 4.46826833021548
9 4.45826398079704
10 4.45922419187449
11 4.47585167826741
12 4.47194870027148
13 4.46723716488802
14 4.47018621823371
15 4.47358833459772
16 4.47195459568176
17 4.47033388903216
18 4.4498546744996
19 4.44664404827475
20 4.44293314810945
21 4.44068307268909
22 4.43313233488225
23 4.42815492096964
24 4.42495205082763
25 4.42430478535892
26 4.42472683341036
27 4.42867936475982
28 4.422987227751
29 4.41119724678434
30 4.41172182587486
31 4.41081636871594
32 4.40910734296638
33 4.4081702362809
34 4.39736363503578
35 4.39422429316931
36 4.3812548111932
37 4.38571175110434
38 4.38478822060051
39 4.38550166669403
40 4.39094375472295
41 4.38755404651025
42 4.37646122267298
43 4.37539041917871
44 4.37264969500076
45 4.37800571983602
46 4.38029715579347
47 4.37919457970844
48 4.39269494356256
49 4.39102692985732
};
\addplot [color0, opacity=0.2]
table {%
0 4.70987197553337
1 4.66451120155818
2 4.6534380359394
3 4.5860585295998
4 4.55080002313681
5 4.5821514497476
6 4.52510754959481
7 4.510982620087
8 4.50002170029678
9 4.49200869942339
10 4.49064440670878
11 4.48440479984976
12 4.48488468085725
13 4.48809357210373
14 4.48316106660342
15 4.49812260699115
16 4.49295832961442
17 4.49039226863188
18 4.48954161721843
19 4.47652910472233
20 4.47884249451836
21 4.47928173420076
22 4.47826457633839
23 4.47795012090111
24 4.42891444487955
25 4.43234876063366
26 4.43223546085422
27 4.43380069545318
28 4.44618069991836
29 4.44779758455101
30 4.45349550649067
31 4.45224855371402
32 4.44886108034571
33 4.45672058412376
34 4.4569963003185
35 4.44171422429075
36 4.44562292980804
37 4.44185645787983
38 4.43287789659144
39 4.43409826821326
40 4.43481861170979
41 4.43319225301942
42 4.43492190773248
43 4.40840598898432
44 4.40603958316068
45 4.34832904335766
46 4.34769179586851
47 4.34660494250318
48 4.34894313456736
49 4.35262760950605
};
\addplot [color0, opacity=0.2]
table {%
0 4.80412326210632
1 4.74317215290151
2 4.5969305680176
3 4.6160495259674
4 4.64656999621316
5 4.57636040645436
6 4.56565350382267
7 4.53322117413774
8 4.51879744413958
9 4.48878054768399
10 4.48627784148659
11 4.49445377992158
12 4.48711755246571
13 4.47433179051363
14 4.47312639804316
15 4.4630565836734
16 4.45635023849965
17 4.44815005613461
18 4.45409321523428
19 4.4596450611507
20 4.44973541262057
21 4.45768167845664
22 4.46141901963025
23 4.47812686129284
24 4.46679307530273
25 4.45559350875503
26 4.44991060948255
27 4.44685856839642
28 4.44318328216828
29 4.44521659814983
30 4.44468559903208
31 4.43229292239692
32 4.43329050411283
33 4.4327902662727
34 4.43643860744169
35 4.4361761355088
36 4.43169711032076
37 4.41777518505826
38 4.41651578171643
39 4.41042321514166
40 4.40815441581095
41 4.41165290319183
42 4.40301963804793
43 4.40177085130112
44 4.39698844984216
45 4.40063545801428
46 4.40384082871014
47 4.39009540576963
48 4.391012770118
49 4.38948856019264
};
\addplot [color0, opacity=0.2]
table {%
0 4.6177970824874
1 4.61204720974516
2 4.61455841483143
3 4.46828208075555
4 4.46276064754261
5 4.41230023693465
6 4.40606175355648
7 4.38838815195228
8 4.3881052299842
9 4.38176533822101
10 4.39641243589137
11 4.39171168093534
12 4.39263004557107
13 4.39677333703418
14 4.39228076886325
15 4.39609653289814
16 4.39469424355672
17 4.37564000143115
18 4.37133474266241
19 4.37978297463707
20 4.37054947634852
21 4.37777687995454
22 4.38293784314729
23 4.37561874586955
24 4.37281933477266
25 4.38121057567155
26 4.38854553939183
27 4.39550702408701
28 4.40265844906377
29 4.40220967235812
30 4.38750849175982
31 4.3921808392224
32 4.39791516678544
33 4.39880456106771
34 4.39706032088832
35 4.39685809360281
36 4.39769024097798
37 4.3961848984518
38 4.39915047635316
39 4.40448551149447
40 4.39002349139292
41 4.3918407226102
42 4.38423096636153
43 4.37995102762487
44 4.37820646715677
45 4.3723277882492
46 4.36819936914275
47 4.3378024357515
48 4.33281400504952
49 4.33410671200274
};
\addplot [color0, line width=1pt]
table {%
0 4.87667918575381
1 4.78965289690865
2 4.68143685977571
3 4.61864918168287
4 4.5811593281322
5 4.5570172371283
6 4.5416473638832
7 4.51671643315148
8 4.50759714743121
9 4.49651708301851
10 4.49563804468222
11 4.49233370258624
12 4.49338937549609
13 4.48618422063349
14 4.48288919859547
15 4.47467165346568
16 4.4647661995615
17 4.45614840357265
18 4.45620125759421
19 4.45333841325017
20 4.44983490236414
21 4.45349580294487
22 4.45252404441953
23 4.45233430135016
24 4.44634095056772
25 4.44509148445874
26 4.44791405136549
27 4.44799964556982
28 4.4476841568569
29 4.44520798110509
30 4.44450809166347
31 4.44133592562095
32 4.44165780786987
33 4.44095709393228
34 4.43951066529261
35 4.43679748904173
36 4.43640154045912
37 4.43285545662493
38 4.43241258267724
39 4.43242760234489
40 4.42993950003929
41 4.43126261266502
42 4.42937547726279
43 4.42567607884147
44 4.42550154120501
45 4.4187768724405
46 4.41553860023024
47 4.41109608042607
48 4.4126821157942
49 4.41198080212997
};
\addplot [color1, opacity=0.2]
table {%
0 4.67615019629679
1 4.7368490099531
2 4.5284345602091
3 4.47285647238933
4 4.36522897356313
5 4.30688259093919
6 4.17333242509059
7 4.11011325643932
8 4.00615203852207
9 4.0077208820553
10 3.98713286418465
11 3.97206865028885
12 3.97542239599473
13 4.00643815797068
14 4.01326738721091
15 3.9892479896874
16 3.98298979351477
17 3.97256579973852
18 3.83065846907533
19 3.83033992336492
20 3.8320433520682
21 3.84662245714527
22 3.84200077579357
23 3.83936031659506
24 3.78359022605657
25 3.51700262628326
26 3.5293657352723
27 3.4879829182661
28 3.4872028925311
29 3.48921444512752
30 3.48974908419932
31 3.51449403819861
32 3.50847170416916
33 3.5084559702298
34 3.50322429630116
35 3.49057989282359
36 3.48704608467409
37 3.47924022223255
38 3.43665724445574
39 3.42226016254764
40 3.41656769739237
41 3.39723827366061
42 3.36930167611708
43 3.22433001088456
44 3.19540107206863
45 3.17749741608747
46 3.19318272810195
47 3.19011968521672
48 3.18814805640032
49 3.19036806859126
};
\addplot [color1, opacity=0.2]
table {%
0 4.68934888837151
1 4.46069841693828
2 4.5670422339022
3 4.4581963054921
4 4.41409403678732
5 4.3328520614306
6 4.23955128883196
7 4.17876251768549
8 4.15320881845818
9 3.99863817069951
10 4.00818151491802
11 3.94640020226578
12 3.98403070797133
13 3.97192714114795
14 3.95664191007545
15 3.98194937625619
16 3.94507328842797
17 3.9522571622924
18 3.92196544870713
19 3.92156922431805
20 3.91930639198831
21 3.78775450261046
22 3.77617978487082
23 3.7594640620159
24 3.76964063632507
25 3.70806819733148
26 3.7108064214007
27 3.67256188747943
28 3.67229846570244
29 3.66679431076802
30 3.62508880609895
31 3.62530901869707
32 3.62454853369258
33 3.60026614078604
34 3.60559509892599
35 3.44685848243243
36 3.42960192710914
37 3.40457489577551
38 3.40517737789996
39 3.40952502698731
40 3.40486946157355
41 3.39386160205317
42 3.39145722454463
43 3.38761122456323
44 3.39431327979516
45 3.39319053306229
46 3.39913738643296
47 3.31149754779797
48 3.2322733812026
49 3.22911105401915
};
\addplot [color1, opacity=0.2]
table {%
0 4.67500545334647
1 4.62787984264421
2 4.57994233687929
3 4.50894360499706
4 4.45180444050187
5 4.36628451956015
6 4.28626952789825
7 4.28134076816469
8 4.17588239492985
9 4.00917403102096
10 3.98968839265903
11 3.97453176038168
12 3.99506879351029
13 3.98874189528163
14 3.99330142537268
15 3.95141283371855
16 3.94919253529104
17 3.94899642770701
18 3.94792974468412
19 3.94424637664115
20 3.94682772906523
21 3.94095392814065
22 3.94498877529548
23 3.94597231931235
24 3.94461349169585
25 3.94185398981906
26 3.86484807000583
27 3.88126755837
28 3.81236981212327
29 3.79583933768902
30 3.6821566321075
31 3.6840094414487
32 3.65606345922424
33 3.64125459767737
34 3.60804682913607
35 3.60253602169641
36 3.59701497871808
37 3.56868201423001
38 3.48889819285611
39 3.49024104160207
40 3.35518120919602
41 3.3456565462271
42 3.34510378708763
43 3.35070921671204
44 3.34657237067917
45 3.32295158648559
46 3.30366674947037
47 3.30690658991719
48 3.30680819109973
49 3.29836057820968
};
\addplot [color1, opacity=0.2]
table {%
0 4.8922510333268
1 4.76701867865978
2 4.63158062026126
3 4.6126973099421
4 4.5476212211926
5 4.2344124763169
6 4.12183964766951
7 4.03880933337668
8 3.81899303455907
9 3.81945541552916
10 3.7345224384875
11 3.74255042906862
12 3.74243475724779
13 3.71352402294032
14 3.72369256289624
15 3.72500940875759
16 3.68934512672198
17 3.68665711344342
18 3.65286377340605
19 3.63438418004589
20 3.56279433313751
21 3.42793808737514
22 3.41200967851581
23 3.39854967874459
24 3.33998045858558
25 3.30104832463954
26 3.28960967064691
27 3.28094661546048
28 3.28759939885581
29 3.17199580397712
30 3.16955075183711
31 3.17330154365472
32 3.18552253569816
33 3.13017969270964
34 3.13308066248895
35 3.09145568369671
36 3.08229523703911
37 3.08101230020176
38 3.05561008788604
39 3.07497912908726
40 3.05913181796541
41 3.04698687337146
42 3.04762388650548
43 3.04768082299153
44 3.06128422892741
45 3.05934161903141
46 3.05720299279684
47 3.01700082743
48 3.01408427987473
49 3.02791682799041
};
\addplot [color1, opacity=0.2]
table {%
0 5.3061953297237
1 4.916026174574
2 4.40393026331058
3 4.41244573330302
4 4.41600464745605
5 4.218228807239
6 4.07561235919022
7 3.9954223821656
8 3.96125589427631
9 3.95632860059578
10 3.94614855976889
11 3.95396480137689
12 3.93563381506497
13 3.759996947167
14 3.75975302053084
15 3.75856604253977
16 3.76089751513702
17 3.75391687105885
18 3.74273639713925
19 3.75635613606475
20 3.56300065228615
21 3.56841406347293
22 3.56431151515905
23 3.56300352176508
24 3.55744458109912
25 3.57769948270347
26 3.57016217164169
27 3.57228196320179
28 3.4761684574869
29 3.4788250916237
30 3.47310652442517
31 3.46963021958562
32 3.40319321170862
33 3.36679709988912
34 3.36791521438135
35 3.29987230066797
36 3.29799386833933
37 3.29717342849449
38 3.28647263586507
39 3.28345013952724
40 3.28107193679276
41 3.28200195074879
42 3.25467974087001
43 3.25759195457809
44 3.24753488256603
45 3.24698679443735
46 3.24286081017556
47 3.23882392176644
48 3.22981385107889
49 3.2168572982469
};
\addplot [color1, opacity=0.2]
table {%
0 4.87232421768655
1 4.67886176664149
2 4.58902883326912
3 4.52957112143758
4 4.42853986646615
5 4.42655973179069
6 4.31619515815174
7 4.04492826155641
8 3.9933592003989
9 3.94621899971184
10 3.93630446998405
11 3.92327508726567
12 3.92495440083017
13 3.85854865688064
14 3.85063553441183
15 3.76449104507074
16 3.73825852241754
17 3.70980362538969
18 3.54079963013662
19 3.53780543993365
20 3.53140782139032
21 3.53336778379642
22 3.53352375243968
23 3.47601673329318
24 3.4590522427281
25 3.4603394379015
26 3.34534040236065
27 3.35071256015111
28 3.35103565823405
29 3.34995854381092
30 3.34329104166686
31 3.33390351123575
32 3.24303506714525
33 3.2357810188251
34 3.2524023734845
35 3.25598530858184
36 3.27514582880831
37 3.2736188790822
38 3.27465866812758
39 3.27969455527477
40 3.2779056832478
41 3.24892259808214
42 3.25376997582355
43 3.23781427914976
44 3.13725060666988
45 3.15228387524109
46 3.15628201237525
47 3.15746569306093
48 3.16083284004508
49 3.16101940481762
};
\addplot [color1, opacity=0.2]
table {%
0 4.81694912973668
1 4.35492794083866
2 4.38621904981506
3 4.31184163638344
4 4.21377729977226
5 4.25763154417009
6 4.25527896275083
7 4.21685358519638
8 4.1923409617283
9 4.16095264861795
10 4.14828657702018
11 4.06147249835139
12 4.0533011593701
13 4.05435810703375
14 3.86445253269882
15 3.84439792392179
16 3.83432298122615
17 3.81502571503889
18 3.81935344328943
19 3.81089188884068
20 3.79652634080939
21 3.78764283080562
22 3.77903045632151
23 3.75819354496241
24 3.72399783959571
25 3.70161560362608
26 3.64123962150204
27 3.57317956325671
28 3.45843692619961
29 3.41246803356565
30 3.39762688640504
31 3.39909359829445
32 3.39929912583461
33 3.40005031524888
34 3.3924035189187
35 3.39684116231527
36 3.38864414024315
37 3.34934498141996
38 3.35016816032549
39 3.35354035282755
40 3.35215666092992
41 3.32117460931491
42 3.31657139761103
43 3.31240808994019
44 3.29592772677476
45 3.27012891775532
46 3.27488343025392
47 3.27079156445508
48 3.26579127881059
49 3.25279675156108
};
\addplot [color1, opacity=0.2]
table {%
0 4.94570365906482
1 4.70938810208295
2 4.71155154534181
3 4.51913470135649
4 4.46034641819229
5 4.26501138047003
6 4.24982195608395
7 4.21026221958765
8 4.1502299247372
9 4.16277610688062
10 4.14867624576796
11 4.06117620727327
12 4.01728536165136
13 3.87532617927224
14 3.87929713265939
15 3.67210728975721
16 3.66866621012381
17 3.6989982054489
18 3.70734280884763
19 3.70135669377905
20 3.63226801111934
21 3.6277761909579
22 3.61445578104125
23 3.57125761836152
24 3.52967441741915
25 3.51838508856003
26 3.46766787136264
27 3.47869887425449
28 3.46002191041283
29 3.38916161999816
30 3.36248594811384
31 3.34776507734814
32 3.3454064784982
33 3.36850122680307
34 3.36919867311571
35 3.36656235242115
36 3.19204516028731
37 3.20367964682039
38 3.20535868276352
39 3.21092876519814
40 3.21525686497018
41 3.20511782622111
42 3.20019833364082
43 3.19539199835677
44 3.19410309091095
45 3.20050006964175
46 3.20503650553199
47 3.20550391776828
48 3.20468565819185
49 3.19473305656989
};
\addplot [color1, opacity=0.2]
table {%
0 4.93842673799031
1 4.35331654202991
2 4.22250393196854
3 4.21116373971607
4 4.24975268585445
5 4.173346365522
6 4.15400993212057
7 4.03320758261315
8 3.77352111945226
9 3.75998552189273
10 3.75881865279792
11 3.71370122252323
12 3.65781270757354
13 3.60210214784001
14 3.58673456658602
15 3.52510071826115
16 3.49785355499093
17 3.48582425447157
18 3.48599430057477
19 3.46423472447769
20 3.40711046250699
21 3.42851889723797
22 3.42523434429746
23 3.41312487669169
24 3.41335118829371
25 3.40696453312801
26 3.3735944025391
27 3.36414899876423
28 3.36769814752796
29 3.36338828631521
30 3.32041582115986
31 3.32693163633432
32 3.3189747136732
33 3.33190675432185
34 3.2784885353969
35 3.2215077622915
36 3.2252499418722
37 3.14803359391669
38 3.14107278677892
39 3.07723169998764
40 3.07971180803999
41 3.00948364411275
42 3.017594896329
43 2.99680362091888
44 2.99628579145762
45 2.99373059551842
46 2.994431502876
47 2.99441950183284
48 2.99941695074599
49 2.9978133458227
};
\addplot [color1, opacity=0.2]
table {%
0 4.70098750937017
1 4.62410475222662
2 4.67379803947043
3 4.31247885026662
4 4.21920558668766
5 4.1833957330441
6 4.19196128738491
7 4.18572730469097
8 4.18702799409443
9 4.17019435677085
10 4.19124710022927
11 4.09987096422082
12 4.07440921870852
13 4.08742629306566
14 4.07265213525953
15 4.06938612225049
16 3.99198611008097
17 3.99481414219767
18 3.99061431070117
19 3.94826966556536
20 3.9579470814872
21 3.93057776919983
22 3.93123346473817
23 3.88272915181335
24 3.89409015927405
25 3.89824500003655
26 3.89696374865245
27 3.8298067196953
28 3.81291720154444
29 3.80656945258189
30 3.80783785445702
31 3.81430040313045
32 3.71670145610634
33 3.71458709337424
34 3.7025395060595
35 3.69919727975844
36 3.69935134412555
37 3.68613469481183
38 3.68219493344201
39 3.67837738596763
40 3.62331804612835
41 3.62328973796858
42 3.61672935500121
43 3.61692470631904
44 3.61600438972609
45 3.60444972578122
46 3.61757730470998
47 3.61077954734025
48 3.59547711556082
49 3.56766540701033
};
\addplot [color1, opacity=0.2]
table {%
0 6.14996748517936
1 5.06621016299687
2 4.6506327304609
3 4.61989775318083
4 4.49273252080008
5 4.2622765454955
6 4.21345958250417
7 4.19401043956283
8 4.17444872849405
9 4.06185812876612
10 4.05775990048257
11 4.02756609249435
12 4.02552199839531
13 3.95864439644807
14 3.95749377185111
15 3.95518758243978
16 3.9452568393417
17 3.75507548855733
18 3.7689965206211
19 3.79418250414961
20 3.53121743839707
21 3.53841515348759
22 3.51784861803079
23 3.49902400790255
24 3.4563253475332
25 3.45702745489736
26 3.45541865777401
27 3.46028197546394
28 3.43893901648328
29 3.41829321172591
30 3.36173208894091
31 3.33784551161171
32 3.35095770740015
33 3.34723981258837
34 3.36334323135214
35 3.34890087158334
36 3.3528230206144
37 3.35162569190851
38 3.37924746237374
39 3.35549308750245
40 3.36856620998382
41 3.38383145854144
42 3.37902440966069
43 3.3685402399155
44 3.36950121632208
45 3.35009509063435
46 3.34833414337166
47 3.34669162441702
48 3.30087096468891
49 3.29628412555624
};
\addplot [color1, opacity=0.2]
table {%
0 4.64350863619409
1 4.60137164654908
2 4.64597345293866
3 4.58210602130238
4 4.55081942347809
5 4.2815387656898
6 4.27586588284945
7 4.27257312098492
8 4.20190734855382
9 4.06798039112019
10 4.06238774374612
11 4.06922494507691
12 3.94927332017041
13 3.86973891555761
14 3.82626885869379
15 3.67966064800716
16 3.67609227070972
17 3.56508136691895
18 3.58847237647747
19 3.59198626921428
20 3.59878935129112
21 3.59984405286232
22 3.60532059713181
23 3.59456631300412
24 3.57816662834176
25 3.60745121551793
26 3.60348676915208
27 3.59646470144108
28 3.5952523575319
29 3.59673945149248
30 3.59216089870375
31 3.59568785796461
32 3.59061474949913
33 3.60267682515755
34 3.578693870003
35 3.57829684161156
36 3.57841528244029
37 3.5693742117993
38 3.56097094854504
39 3.55866335839249
40 3.53005089808479
41 3.53164948318077
42 3.51653936801223
43 3.48204804912585
44 3.4760872700783
45 3.475505418267
46 3.44163662368011
47 3.44262543153464
48 3.37921892059705
49 3.37786415243508
};
\addplot [color1, line width=1pt]
table {%
0 4.94223485635727
1 4.65805441967791
2 4.54921979981891
3 4.46261110414725
4 4.40082726006266
5 4.275701710139
6 4.21276650087718
7 4.14683423100201
8 4.06569395485037
9 4.01010693780508
10 3.99742953833718
11 3.96215023838229
12 3.94459571970738
13 3.89556440505046
14 3.87368256985389
15 3.82637641505565
16 3.80666122899863
17 3.7782513476886
18 3.74981060197167
19 3.74463525219959
20 3.68993658046223
21 3.66815214309101
22 3.66217812863628
23 3.64177184537182
24 3.62082726807899
25 3.59130841287035
26 3.56237529519253
27 3.54569452798372
28 3.51832835371947
29 3.4949372990563
30 3.46876686150961
31 3.46852265479201
32 3.4452323952208
33 3.43730804563425
34 3.429577650797
35 3.39988282999002
36 3.38380223452258
37 3.36770788005777
38 3.35554059844327
39 3.34953205874185
40 3.33031569119208
41 3.31576788362357
42 3.30904950426695
43 3.28982118445462
44 3.27752216049801
45 3.27055513682861
46 3.26951934914805
47 3.25771882104478
48 3.23978512402471
49 3.23423250590253
};
\addplot [color2, opacity=0.2]
table {%
0 5.1118343200149
1 4.56924821970868
2 4.4238797727561
3 4.33436421013521
4 4.27795154727225
5 4.26656478512953
6 4.07654244233884
7 4.03867592762536
8 3.8913480273791
9 3.86543934137464
10 3.78135147504829
11 3.69359758587629
12 3.64311631012195
13 3.59254602155337
14 3.58887060230855
15 3.56598260708091
16 3.4583613088207
17 3.44040537135958
18 3.3420230652001
19 3.28868133436612
20 3.27253103237092
21 3.26027657507689
22 3.22811671422154
23 3.24229816409497
24 3.27474801766244
25 3.27968560967002
26 3.22153589566385
27 3.21760060909654
28 3.2181837644491
29 3.18247173317957
30 3.13184631534172
31 3.11581289941616
32 3.01554338676475
33 3.0178485197273
34 2.99724307357033
35 2.99211657550053
36 2.98381620015323
37 2.98849720059437
38 2.99487433279445
39 2.98572351140045
40 2.99371814575652
41 2.98995868033085
42 2.8977058718907
43 2.87055750953887
44 2.86872393378226
45 2.86257544704968
46 2.82027328725872
47 2.82870864676646
48 2.87943013599616
49 2.87228462601968
};
\addplot [color2, opacity=0.2]
table {%
0 4.76988494191594
1 4.58392158047101
2 4.59255766329565
3 4.31109563078608
4 4.25449225226654
5 4.08334214762297
6 3.97754075086401
7 3.95686461548579
8 3.7335065759208
9 3.58757265356408
10 3.57639277667279
11 3.49455391703427
12 3.45947302214023
13 3.4113339839476
14 3.30734073753318
15 3.15699752649015
16 3.08544443407475
17 3.09591216152055
18 3.09375283451606
19 3.0642970206313
20 3.06255716109442
21 3.04524254369712
22 3.02651750925419
23 3.01231355107313
24 2.94320671572412
25 2.9269224575103
26 2.92962499748985
27 2.92787546709814
28 2.92607125564449
29 2.94316678558675
30 2.94826894931312
31 2.91548011347298
32 2.91495501969826
33 2.89919308172685
34 2.89662664338969
35 2.89924152485343
36 2.88497463538649
37 2.88504784422636
38 2.88401948843054
39 2.88086827728758
40 2.8567008489281
41 2.85239069976279
42 2.84305816585768
43 2.84339432216265
44 2.84207480199681
45 2.84200445435612
46 2.83854046575806
47 2.83847731818534
48 2.83364427731149
49 2.82834540332894
};
\addplot [color2, opacity=0.2]
table {%
0 4.70677530840991
1 4.55946767361493
2 4.40247113370245
3 4.30369124400945
4 4.20105399841214
5 4.16323861029269
6 4.14624321619224
7 4.13855072724835
8 3.9682366339719
9 3.96189965187329
10 3.90423053959575
11 3.81745866951256
12 3.8294434404899
13 3.78553746781853
14 3.76821612116663
15 3.68631673778803
16 3.47987756693743
17 3.43854912145345
18 3.39559866503164
19 3.23341678699828
20 3.25218751507213
21 3.21809178316374
22 3.19912864851691
23 3.17883586962567
24 3.16608843202115
25 3.15881657630875
26 3.04714840279926
27 3.04192042932364
28 3.01810824918793
29 3.0092140450165
30 2.99730301048934
31 2.92211956437663
32 2.88164910080305
33 2.87861976111423
34 2.84135325953868
35 2.86300803858626
36 2.85661902793563
37 2.85699970224498
38 2.86414895980017
39 2.86670597180461
40 2.82403570926865
41 2.81993046103728
42 2.81192677253235
43 2.8149572811693
44 2.80382605340928
45 2.79870383164824
46 2.80044920621224
47 2.78079511158789
48 2.78024818181542
49 2.75222311696959
};
\addplot [color2, opacity=0.2]
table {%
0 4.94069812964728
1 4.64515950368429
2 4.46666594836409
3 4.15300784116811
4 4.12785388698031
5 4.11418871325337
6 4.0471642700045
7 4.0563930560101
8 4.04654439787507
9 3.86476620713865
10 3.80859681539502
11 3.74895231934596
12 3.69416959508613
13 3.66482949777212
14 3.63160895667187
15 3.63573145958826
16 3.58989044185409
17 3.46173213568603
18 3.42805354579471
19 3.35982617397271
20 3.36625057832443
21 3.38399770771384
22 3.32947913035948
23 3.32716645800653
24 3.32309556899131
25 3.29174202012734
26 3.28853433265538
27 3.23342502912079
28 3.16824708247397
29 3.13986107380184
30 3.10754235881135
31 3.07466329925909
32 3.10487325664785
33 3.08064572198479
34 3.05047644279675
35 3.05445290672915
36 3.04817752555004
37 3.04827545242172
38 3.01756257102695
39 2.96916175420444
40 2.94202321211836
41 2.91817824911681
42 2.8801924152318
43 2.87959995332137
44 2.8776818044733
45 2.88929153116687
46 2.89654039300837
47 2.90573921073158
48 2.91035674594809
49 2.90811131220345
};
\addplot [color2, opacity=0.2]
table {%
0 4.82748806895051
1 4.43168960905861
2 4.34567831445867
3 4.19694517743494
4 4.16827832892155
5 4.00587969713178
6 3.99162769701827
7 3.96678556853303
8 3.91728100957524
9 3.82213870058513
10 3.79930422742443
11 3.67915319125974
12 3.6570301426664
13 3.52812013673678
14 3.52791504111172
15 3.44891668562767
16 3.34889515740034
17 3.33093821112166
18 3.33058724129988
19 3.23805913585819
20 3.1790052218254
21 3.15210988358962
22 3.12861239352068
23 3.12536186394331
24 3.09395558918016
25 3.08253302582656
26 3.07563509836479
27 3.06867653311026
28 3.06810425477471
29 3.06835936412267
30 3.0447077134413
31 3.0375333150218
32 3.04662135553427
33 3.04645724788745
34 2.96633084370761
35 2.96519914014171
36 2.96002898556435
37 2.94870551456747
38 2.94818665228677
39 2.94794724067577
40 2.93375479115783
41 2.94367903022994
42 2.93981994881596
43 2.94270087810025
44 2.9345338751769
45 2.87262692885802
46 2.87298497746683
47 2.8658912644192
48 2.86702463531917
49 2.86082126699862
};
\addplot [color2, opacity=0.2]
table {%
0 5.09331931736675
1 4.37984456531017
2 4.31634207124117
3 4.16932434421466
4 4.04426460525925
5 4.05213098989818
6 4.0210730306204
7 3.86951923142988
8 3.78620717768599
9 3.77530206545968
10 3.80519909055235
11 3.64237887170212
12 3.64785381662259
13 3.6659909558673
14 3.58792402691932
15 3.46588346947828
16 3.40167370139569
17 3.39939914290477
18 3.38578875128735
19 3.29868054226661
20 3.27529704181228
21 3.28939573744726
22 3.27681408374649
23 3.13835375978358
24 3.11468884789246
25 3.01297462955768
26 3.02499391691218
27 3.01911421169613
28 3.01633079674228
29 3.01268755406597
30 2.99963298118449
31 2.99779033778012
32 2.98562175292866
33 2.96368694894887
34 2.96370866160517
35 2.96615266968755
36 2.9573997080887
37 2.92940851050457
38 2.92096432735495
39 2.87140298344625
40 2.87181043403605
41 2.86663699785315
42 2.86656481559752
43 2.85346843234243
44 2.83618531084182
45 2.82676273320647
46 2.82075876083335
47 2.77853293910396
48 2.77003334100214
49 2.72890556586448
};
\addplot [color2, opacity=0.2]
table {%
0 4.70762266100452
1 4.40396107679409
2 4.37124297668904
3 4.28464766611427
4 4.1825160131395
5 4.2002695474441
6 4.22589063453736
7 4.03855199131843
8 4.0570811801294
9 3.94165040655997
10 3.83757127591966
11 3.73772503179373
12 3.73749643011155
13 3.7172520684799
14 3.53009762480085
15 3.51672074069507
16 3.40466616201529
17 3.38400645457208
18 3.40046469929907
19 3.33049255275003
20 3.22798334988576
21 3.21775412220006
22 3.16051762617865
23 3.14117421587874
24 3.11764712811964
25 3.09805873716332
26 3.09785095435563
27 3.10007010739203
28 3.10121694447088
29 3.10640013828382
30 3.10398820529875
31 3.0854358641086
32 3.08895199525354
33 3.03448237214162
34 2.97779111756132
35 2.96078034053145
36 2.96133439055849
37 2.96080686931395
38 2.92977216896909
39 2.92406285920497
40 2.91475774889231
41 2.91722496209701
42 2.90661514135608
43 2.89471878200585
44 2.81154390981504
45 2.81726971995018
46 2.80852591416346
47 2.70729954595159
48 2.71309487951303
49 2.7136961998979
};
\addplot [color2, opacity=0.2]
table {%
0 4.99083581435396
1 4.67584558017865
2 4.54907243281237
3 4.44886576031775
4 4.41467350305322
5 4.11258628057117
6 4.06306110719226
7 3.97398447116397
8 3.86367588731861
9 3.74959833224624
10 3.72502305866279
11 3.70281532225979
12 3.68088553037325
13 3.63783138031018
14 3.65328608355177
15 3.5266626419009
16 3.50228738308237
17 3.42093827948922
18 3.32691173865631
19 3.14166031937345
20 3.11821687788653
21 3.12521102294794
22 3.10701782579213
23 3.06339268409577
24 3.06628921676319
25 3.07060520656593
26 3.06890559217646
27 3.03277004121107
28 3.00323418415473
29 2.97727475732946
30 2.96941266538184
31 2.96084032547352
32 2.90660535511577
33 2.90526906498403
34 2.89338419111963
35 2.89542534095207
36 2.87591924117839
37 2.88053142206263
38 2.85709816273425
39 2.85521634698121
40 2.85636071912517
41 2.85518164756868
42 2.89280821467917
43 2.89666742883186
44 2.89243197849676
45 2.89099192156082
46 2.8815559163032
47 2.88397449208345
48 2.87376852283791
49 2.86204637965104
};
\addplot [color2, opacity=0.2]
table {%
0 4.667652514183
1 4.3853144367607
2 4.34031689249997
3 4.13161648128869
4 4.10230186933915
5 4.08322511300376
6 4.03651337360408
7 3.95470385417378
8 3.83338928603415
9 3.73960830148616
10 3.69901491917382
11 3.63664984062163
12 3.60972501610819
13 3.60390756146961
14 3.53353813433449
15 3.52500889588619
16 3.39444149211395
17 3.35191524255101
18 3.32419365966492
19 3.30166288202923
20 3.19615939996995
21 3.17136361957304
22 3.16681017308593
23 3.16054695749341
24 3.16278504345139
25 3.1582584703725
26 3.10717390490612
27 3.04800200354932
28 3.03334407414406
29 3.02117562437389
30 2.97940400093622
31 3.00425718606867
32 2.96437582105029
33 2.9621775392615
34 2.95866159161218
35 2.97144743678627
36 2.95764569935293
37 2.88319305140855
38 2.89522459129367
39 2.88404226759984
40 2.87873485468789
41 2.8760299946402
42 2.87482450436761
43 2.83784739319936
44 2.81727422962938
45 2.78610808757371
46 2.7746925672449
47 2.74310809188735
48 2.73650645364313
49 2.73428300596812
};
\addplot [color2, opacity=0.2]
table {%
0 4.88496947335345
1 4.49767757541771
2 4.45185898565036
3 4.3944476061588
4 4.39644065150481
5 4.28256716176258
6 4.18053190516939
7 3.93002701846611
8 3.9654604570846
9 3.93666778105365
10 3.92980786560043
11 3.91267376093786
12 3.80585861904688
13 3.75203293633466
14 3.70577800864307
15 3.68509908926376
16 3.56729790427754
17 3.57387519006368
18 3.56251263679749
19 3.48792260684809
20 3.48694992626262
21 3.40892088308454
22 3.42093643800701
23 3.3870310458973
24 3.37101750454729
25 3.20910804981604
26 3.19976931338957
27 3.2043330302907
28 3.21201896427627
29 3.20994465024537
30 3.07891241702454
31 3.06385913369912
32 3.05459770208236
33 3.0538345887975
34 3.00357945031337
35 2.99610076978317
36 2.98439543756831
37 2.9228448093226
38 2.93385711556797
39 2.93975672079773
40 2.92840683659673
41 2.92944847652656
42 2.92471602119501
43 2.92079369202071
44 2.90526695139067
45 2.89796161126144
46 2.89936186801987
47 2.8799996848117
48 2.87084947873657
49 2.87242293491172
};
\addplot [color2, opacity=0.2]
table {%
0 5.17280289114466
1 4.59688738874111
2 4.47145545724236
3 4.33510733673275
4 4.28472526098116
5 4.28810670503347
6 4.30342576245317
7 4.29103929340039
8 4.17309652528044
9 4.11351120141093
10 4.06754964687578
11 4.01305605542984
12 3.75142023284222
13 3.63764036279319
14 3.60736773428691
15 3.46127812420347
16 3.45996972694823
17 3.45885422874961
18 3.30991257069029
19 3.28321423956511
20 3.2446378532639
21 3.22933374837474
22 3.22994681407394
23 3.1972163931572
24 3.2014905026908
25 3.17159157662783
26 3.147851703317
27 3.01576909446326
28 3.02177544743153
29 3.02298592290703
30 3.01759513314033
31 2.97459325085047
32 2.97656953069707
33 2.98558823620348
34 2.97747043788699
35 2.96419357608176
36 2.93748587859518
37 2.90459503320382
38 2.85339208632262
39 2.86217805294195
40 2.86110056414706
41 2.85540991347393
42 2.84977574494885
43 2.84704277818495
44 2.81204054961027
45 2.80912726941025
46 2.8110705669637
47 2.79844607783725
48 2.79601013466473
49 2.79080798638887
};
\addplot [color2, opacity=0.2]
table {%
0 4.63547432198913
1 4.3901392133301
2 4.36452351420475
3 4.20553649459346
4 4.13159806047389
5 4.1559528092674
6 4.13105367965732
7 3.98279969250797
8 3.85403748849322
9 3.74199854193854
10 3.67352174118446
11 3.62232097862468
12 3.61534501703317
13 3.59852687594248
14 3.51977562829269
15 3.42258540513908
16 3.4051187960582
17 3.37136092369614
18 3.31303111577397
19 3.32535676993067
20 3.22470793451473
21 3.20170321660668
22 3.19302423057465
23 3.15230890938163
24 3.12978064080496
25 3.11801178529913
26 3.10149394517252
27 3.10765168427842
28 3.10118722825237
29 3.10075273561195
30 3.0912276771523
31 3.08829955779165
32 3.06374110234728
33 3.05465192561754
34 3.03785196624411
35 3.03858898037464
36 3.03378743301891
37 3.01930362063219
38 3.0197559574414
39 3.01773544298873
40 3.0216875120908
41 3.0286322794024
42 3.03519055694518
43 3.02456049063181
44 3.02978585551238
45 3.01698941401749
46 3.01739486643436
47 2.93790787140683
48 2.89623596648266
49 2.89597049807463
};
\addplot [color2, line width=1pt]
table {%
0 4.87577981352783
1 4.50992970192251
2 4.42467209690975
3 4.27238748274618
4 4.21551249813365
5 4.15067104670092
6 4.10005565580432
7 4.01649128728026
8 3.92415538722904
9 3.84167943205758
10 3.8006302860088
11 3.72511129536654
12 3.67765143105354
13 3.63296243741881
14 3.58014322496842
15 3.50809861526181
16 3.42482700624822
17 3.39399053859731
18 3.35106921033431
19 3.27943919704915
20 3.24220699102359
21 3.22528340362296
22 3.2055767989443
23 3.17716665603594
24 3.16373276732074
25 3.13152567873712
26 3.10920983810022
27 3.08476735338586
28 3.07398518716686
29 3.0661911987104
30 3.03915345229294
31 3.0200570706099
32 3.00034211491026
33 2.99020458403293
34 2.96370647327882
35 2.96389227500067
36 2.95346534691255
37 2.93568408587527
38 2.92657136783524
39 2.91706678577779
40 2.90692428140046
41 2.90439178266997
42 2.89359984778483
43 2.88552574512579
44 2.86928077117791
45 2.85920107917161
46 2.85351239913892
47 2.82907335456438
48 2.82726689610588
49 2.81832652468975
};
\node[text=color0,anchor=center,rotate=-0] 
at (axis cs:27,4.6) {observe only};

\node[text=color1,anchor=center,rotate=-20]%,fill=white,fill opacity=0.7,text opacity=1] 
at (axis cs:30,3.7) {random intervention};

\node[text=color2,anchor=center,rotate=-20] 
at (axis cs:22,2.9) {\bf{Algorithm 1}};
\end{axis}
\end{tikzpicture}